\documentclass[10pt,twocolumn,letterpaper]{article}

\usepackage[pagenumbers]{cvpr}      
\usepackage{iftex}
\ifPDFTeX
  \usepackage{CJKutf8}
  \newcommand{\promptchinese}[1]{\begin{CJK*}{UTF8}{gbsn}#1\end{CJK*}}
\else
  \usepackage[no-math]{fontspec}
  \IfFontExistsTF{Songti SC}{\newfontfamily\promptzh{Songti SC}}{%
    \IfFontExistsTF{Noto Serif CJK SC}{\newfontfamily\promptzh{Noto Serif CJK SC}}{%
      \newfontfamily\promptzh{FandolSong-Regular.otf}}}
  \newcommand{\promptchinese}[1]{{\promptzh\ifXeTeX\XeTeXlinebreaklocale "zh"\relax\fi #1}}
\fi
\definecolor{cvprblue}{rgb}{0.21,0.49,0.74}
\usepackage[pagebackref,breaklinks,colorlinks,allcolors=cvprblue]{hyperref}
\usepackage{multirow}
\usepackage{colortbl} 
\usepackage{tabularx}

\def\confName{CVPR}
\def\confYear{2027}
\newtheorem{lemma}{Lemma}[section]
\newtheorem{corollary}[lemma]{Corollary}
\title{Decoupled Visual On-Policy Self-Distillation}
\author{%
  \parbox{\textwidth}{%
    \centering
    \textbf{Baoquan Gong}\textsuperscript{1},
    \textbf{Hexu Ren}\textsuperscript{2},
    \textbf{Xinrui Jiang}\textsuperscript{1},
    \textbf{Bing Cao}\textsuperscript{1,\ensuremath{\dagger}}\\[4pt]
    \textsuperscript{1}Tianjin University, Tianjin, China\\
    \textsuperscript{2}University of Electronic Science and Technology of China, Chengdu, China
  }%
}
\begin{document}
\maketitle
\begingroup
  \renewcommand{\thefootnote}{\ensuremath{\dagger}}
  \footnotetext[1]{Corresponding author.}
\endgroup
\begin{abstract}

Visual on-policy self-distillation uses crop-conditioned predictions to improve fine-grained understanding of full images. However, we identify an overlooked limitation termed self-limiting supervision: the student's probability allocation can underweight unresolved candidate discrepancies and lower their alignment penalty through mass changes without correcting them. An exact regional decomposition attributes this limitation under reverse KL to the coupling between student regional mass and conditional alignment. Empirical analysis shows that this coupling weakens supervision at high-disagreement positions, even though they carry much of the useful visual signal. To address these issues, we reformulate regional conditional alignment as a
min--max optimization over the teacher--student mass-calibration interval, yielding \textbf{A}daptive \textbf{D}ecoupled \textbf{D}istillation (ADD). ADD decouples probability-mass calibration from conditional alignment, using adaptive weights to strengthen supervision and holding them fixed during each update to block direct weight-reduction shortcuts, thereby addressing self-limiting supervision. It integrates into existing visual self-distillation methods without additional forward passes. Experiments show that ADD improves existing methods across six fine-grained visual evaluation settings and achieves performance gains when extended to other on-policy distillation tasks.
\end{abstract}    
\section{Introduction}
\label{sec:introduction}

Fine-grained visual question answering often hinges on a local detail in the full image~\cite{khayatkhoei2025mllms}, such as an object's color or material or a single character in small text~\cite{wu2024v,wei2026zooming,fu2026ocrbench}. Vision-OPD~\cite{yuan2026vision} shows that the same vision-language model can answer questions correctly given a relevant crop after failing on the full image. This suggests that the model's fine-grained visual capabilities are underused with full-image inputs. Through visual on-policy self-distillation, Vision-OPD transfers this advantage to the full-image policy by using a crop-conditioned teacher to supervise a full-image student along student-generated trajectories.

\begin{figure*}[t]
    \centering
    \includegraphics[width=\textwidth]{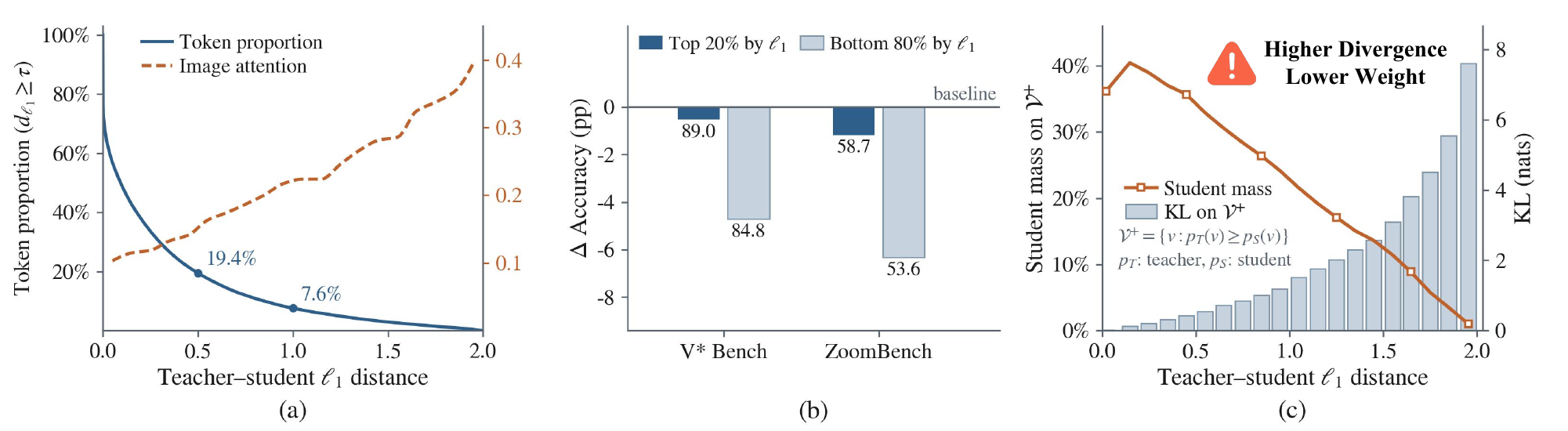}
    \caption{
        \textbf{Self-limiting supervision.}
        (a) Positions with large teacher--student $\ell_1$ distances
        are sparse but generally show greater attention to image tokens.
        (b) Distilling the top 20\% by $\ell_1$ distance nearly matches
        full-sequence distillation and outperforms distilling the remaining
        80\%. Bars show changes from full-sequence accuracy (percentage
        points); labels show accuracy (\%).
        (c) In $\mathcal{V}^{+}=\{v:p_T(v)\geq p_S(v)\}$, conditional KL
        generally rises with disagreement, while the student mass
        weighting it under reverse KL generally falls.
    }
    \label{fig:intro_analysis}
\end{figure*}

Where along a generated response does this supervision matter most? We analyze student trajectories from 256 training examples and measure teacher--student disagreement at each position by the $\ell_1$ distance $d_{\ell_1}=\sum_{v\in\mathcal{V}}|p_T(v)-p_S(v)|$~\cite{wen2023f}. Only 19.4\% of positions have a distance of at least 0.5, and the share of attention assigned to image tokens generally rises with disagreement (Fig.~\ref{fig:intro_analysis}(a)). Distilling only the top 20\% of positions by $\ell_1$ nearly matches full-sequence distillation and outperforms distilling the remaining 80\% by 4.2 and 5.1 percentage points on V$^{*}$Bench~\cite{wu2024v} and ZoomBench~\cite{wei2026zooming}, respectively (Fig.~\ref{fig:intro_analysis}(b)). These results indicate that much of the useful distillation signal is concentrated at a small fraction of high-disagreement positions, consistent with related findings~\cite{liu2026visual,korelaxed}.

However, we identify an overlooked limitation at these informative positions: \emph{self-limiting supervision}. The student's probability allocation can underweight unresolved candidate discrepancies and lower their alignment penalty through mass changes without correcting them. Under reverse KL~\cite{kullback1951information}, Fig.~\ref{fig:intro_analysis}(c) shows that as teacher--student disagreement grows, the conditional KL divergence among candidates favored by the teacher generally increases, while the student mass weighting it generally decreases. Such underweighting can leave visually important discrepancies insufficiently corrected, limiting the transfer of local visual capabilities to the full-image policy.

An exact regional decomposition reveals that, under reverse
KL, \textit{self-limiting supervision} arises from coupling student
regional mass to conditional alignment.
At each generation position, we partition candidates in the vocabulary
by the sign of the teacher--student probability gap:
the positive region contains candidates whose teacher probability
is at least as large as their student probability,
and the negative region contains the remaining candidates.
The objective separates into probability-mass calibration
across regions and conditional-distribution alignment
within regions.
The student's regional mass couples these objectives by serving
as both a calibration variable and an alignment weight:
a small mass suppresses the corrective signal, while gradients
through this weight enable a direct weight-reduction shortcut.
Forward KL and Jensen--Shannon divergence~\cite{lin1991divergence} admit analogous
mass-weighted decompositions.

To address these issues, inspired by robust optimization~\cite{hashimoto2018fairness,sagawa2019distributionally,namkoong2016stochastic}, we reformulate the underweighted regional conditional alignment as a min--max optimization  over the calibration interval bounded by the current teacher and student regional masses, yielding \textbf{A}daptive \textbf{D}ecoupled \textbf{D}istillation (ADD). ADD uses the larger regional mass selected by the inner maximization to strengthen alignment supervision and holds this adaptive weight fixed during each update to decouple mass calibration from conditional alignment and block direct weight-reduction shortcuts. These weights come directly from current predictions, allowing ADD to complement existing visual self-distillation methods without additional forward passes or new conditioning inputs. Our contributions are as follows.

\begin{itemize}
\item We identify \textit{self-limiting supervision} in visual
on-policy self-distillation. Through an exact regional decomposition,
we trace this limitation to the coupling between student regional
mass and conditional alignment.

    \item We propose ADD, which derives adaptive weights through
a robust min--max reformulation and fixes them during updates,
decoupling mass calibration from conditional alignment
to mitigate \textit{self-limiting supervision}.

    \item Experiments show that ADD improves the average performance of existing methods across six fine-grained visual evaluation settings, with gains extending to a broader range of on-policy distillation tasks.
\end{itemize}
\section{Related Work}
\label{sec:rw}

\subsection{Fine-Grained Visual Self-Distillation}

Region-level supervision supports fine-grained visual understanding~\cite{guo2024regiongpt,you2024ferret,yuan2024osprey,chen2023shikra}. ZwZ~\cite{wei2026zooming} uses a strong teacher to generate visual question-answering data from cropped regions for full-image training. Vision-OPD~\cite{yuan2026vision} transfers the model's own crop-conditioned predictions to its full-image policy through on-policy self-distillation. OPD-V~\cite{bi2026opd} selects and weights generation positions by contrasting teacher predictions under zoomed-in and masked views; VAD~\cite{zhang2026vad} reconstructs targets through visual evidence attribution; and EAD~\cite{hu2026evidence} weights teacher corrections by their alignment with the student's responses to visual evidence. These methods refine supervision sources, position selection, or target construction. Even with informative visual guidance, however, the distillation objective can still exhibit self-limiting supervision. ADD complements these methods at the optimization level by adaptively decoupling probability-mass calibration from conditional alignment within each position, strengthening the correction of relative candidate probabilities.

\subsection{Decoupling and Robust Optimization}

In traditional knowledge distillation, DKD~\cite{zhao2022decoupled}
decouples target-versus-rest mass matching from conditional
alignment among non-target classes.
LTKD~\cite{kim2026distilling} groups classes by frequency
and aligns them equally, while
GDKD~\cite{zheng2025rethinking} partitions classes by teacher rank.
These methods replace teacher-mass-dependent weights
with preset coefficients~\cite{luong2026diversity,gong2026multimodal}.
In on-policy setting, training prefixes are student-generated,
regions follow current teacher--student probability gaps,
and reverse KL weights conditional alignment by student regional mass.
The imperfect policy shapes training contexts
and corrective supervision, potentially underweighting
unresolved discrepancies and reducing their penalties through
mass changes without improved alignment.
Large errors in small groups can be obscured in
average risk~\cite{hashimoto2018fairness},
paralleling the suppression of conditional discrepancies
by low regional mass.
Inspired by Group DRO's worst-group
optimization~\cite{sagawa2019distributionally}, ADD minimizes
each region's worst-case weighted conditional loss over
the current teacher--student mass-calibration interval.
Adaptive weights replace original alignment
weights and remain fixed during updates, retaining mass
calibration while reinforcing conditional correction
and blocking direct weight-reduction shortcuts.
\section{Method}

\begin{figure*}
    \centering
    \includegraphics[width=1.0\textwidth]{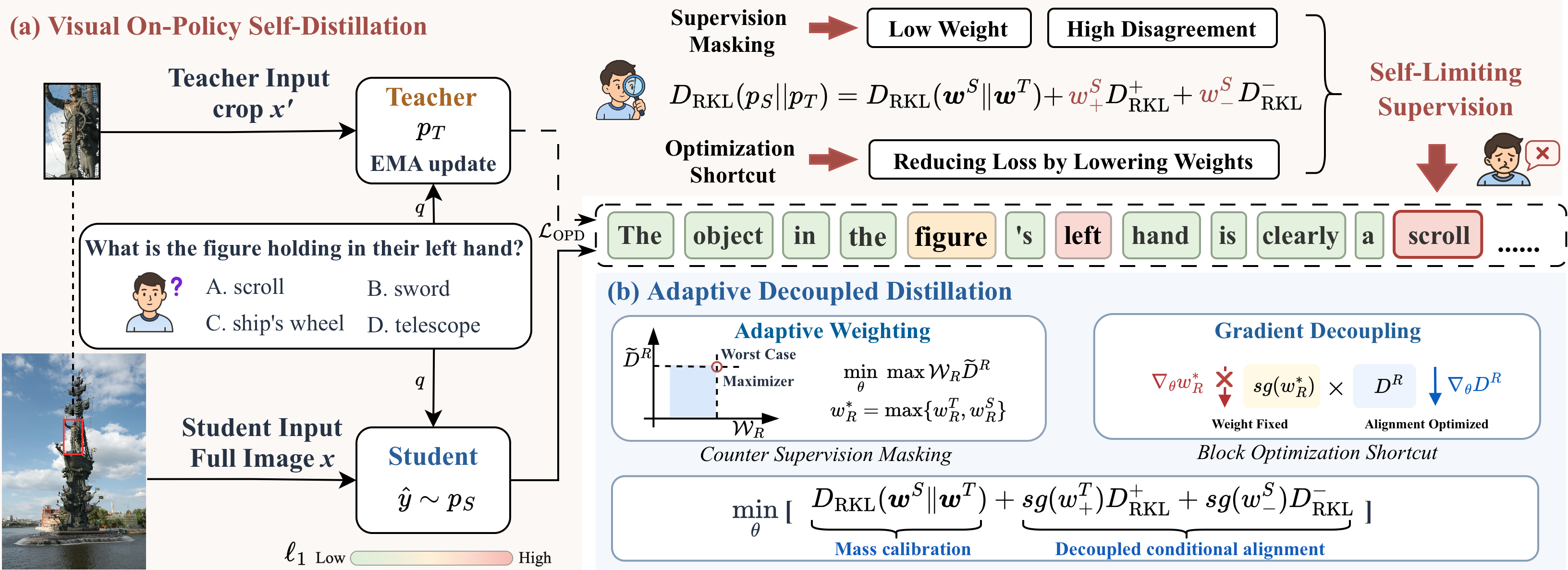}
\caption{
\textbf{Overview}.
(a) A crop-conditioned teacher supervises the full-image student along student-generated trajectories.
Under reverse KL, coupling conditional alignment to student regional
mass can cause supervision masking and optimization shortcuts,
yielding \emph{self-limiting supervision}.
(b) Regional min--max optimization counters masking by minimizing
the worst-case weighted conditional loss; stop-gradient fixes the
resulting weights to block direct weight-reduction shortcuts
while retaining mass calibration.
}
    \label{fig:placeholder} 
\end{figure*}

\subsection{Preliminaries}

Let $\mathcal{D}=\{(x_i,x'_i,q_i)\}_{i=1}^{N}$ be a dataset of full
images $x$, local crops $x'$, and questions $q$. The teacher
$p_T(\cdot\mid x',q)$ and student $p_S(\cdot\mid x,q)$ use the same
model architecture and receive the crop and full image, respectively.
The teacher parameters $\bar\theta$ are updated by an exponential
moving average (EMA) of the student parameters $\theta$~\cite{tarvainen2017mean}, with
gradients stopped through the teacher. Following on-policy
distillation (OPD)~\cite{agarwal2024policy}, the student samples a trajectory
$\hat{y}\sim p_S(\cdot\mid x,q)$ and matches the teacher's token-level
predictions at each shared generation prefix.
\begin{multline}
\mathcal{L}_{\mathrm{OPD}}(\theta)
=\mathbb{E}_{\substack{(x,x',q)\sim\mathcal{D}\\
\hat{y}\sim p_S(\cdot\mid x,q)}}
\!\Bigg[\frac{1}{|\hat{y}|}\sum_{t=1}^{|\hat{y}|}\\
D\Big(p_T(\cdot\mid \hat{y}_{<t},x',q)
\,\Big\|\,p_S(\cdot\mid \hat{y}_{<t},x,q)\Big)\Bigg],
\label{eq:opd}
\end{multline}
where $D$ denotes a distribution divergence. Reverse KL (RKL),
widely used in OPD~\cite{agarwal2024policy,gu2024minillm},
is denoted by $D_{\mathrm{RKL}}$.

\subsection{Self-Limiting Supervision}
\label{sec:region_decomposition}

\paragraph{Region decomposition.}
Fine-grained visual understanding relies on local details to
distinguish similar candidates. In visual self-distillation,
teacher--student probability gaps indicate which candidates to
promote or suppress~\cite{xia2026distillation}, while the teacher's relative
preferences within each set convey finer distinctions~\cite{hinton2015distilling}. To isolate
these two signals, we adopt a decoupled view of distribution
matching~\cite{zhao2022decoupled} and separate mass calibration across candidate
sets from conditional alignment within them.

At each generation position, we partition the full-vocabulary $\mathcal{V}$ according
to the sign of the teacher--student probability gap.
\begin{equation}
\begin{aligned}
\mathcal{V}^{+}
&=\{v\in\mathcal{V}:p_T(v)\geq p_S(v)\},\\
\mathcal{V}^{-}
&=\{v\in\mathcal{V}:p_T(v)<p_S(v)\}.
\end{aligned}
\label{eq:region_partition}
\end{equation}
The positive region contains teacher-favored candidates and ties,
while the negative region contains candidates overestimated by the
student relative to the teacher.

We illustrate the decomposition with RKL. For each region
$R\in\{+,-\}$, the teacher and student masses are
\begin{equation}
w_R^T=\sum_{v\in\mathcal{V}^{R}}p_T(v),
\qquad
w_R^S=\sum_{v\in\mathcal{V}^{R}}p_S(v).
\label{eq:region_mass}
\end{equation}
Normalizing within each region gives the conditional
distributions over its candidates.
\begin{equation}
p_T^R(v)=\frac{p_T(v)}{w_R^T},
\qquad
p_S^R(v)=\frac{p_S(v)}{w_R^S},
\quad v\in\mathcal{V}^{R}.
\label{eq:region_conditional}
\end{equation}
Substituting these expressions into RKL yields~\cite{cover1991elements}
\begin{equation}
\begin{aligned}
D_{\mathrm{RKL}}
&=\sum_R\sum_{v\in\mathcal{V}^{R}}
p_S(v)\log\frac{p_S(v)}{p_T(v)}\\
&=D_{\mathrm{RKL}}(\boldsymbol{w}^{S}\|\boldsymbol{w}^{T})
+\sum_R w_R^S D_{\mathrm{RKL}}^R,
\end{aligned}
\label{eq:rkl_decomposition}
\end{equation}
where $\boldsymbol{w}^{a}=(w_+^a,w_-^a)$ for $a\in\{T,S\}$ and
$D_{\mathrm{RKL}}^R=\mathrm{KL}(p_S^R\|p_T^R)$. The first term
calibrates regional masses, and the second aligns relative
probabilities within regions.

Let $\mathcal{L}_{\mathrm{mass}}
= D_{RKL}(\boldsymbol{w}^{S}\|\boldsymbol{w}^{T})$.
The per-position distillation objective can then be written
equivalently as
\begin{equation}
\min_{\theta}\;
\mathcal{L}_{\mathrm{mass}}
+\sum_{R\in\{+,-\}}w_R^S D^R_{RKL},
\label{eq:regional_objective}
\end{equation}
where $D^R_{RKL}$ denotes the conditional divergence within region $R$
and $w_R^S$ is its weight determined by regional probability mass.

\paragraph{Coupled alignment.}
\label{coupledalign}
In Eq.~\eqref{eq:regional_objective}, the student's regional mass
serves as both a calibration variable and an outer alignment weight,
coupling the two terms through the same variable.
Fixing the sampled prefix, teacher distribution, and regional
partition gives
\begin{equation}
\nabla_\theta\!\left(w_R^S D_{\mathrm{RKL}}^R\right)
=
w_R^S\nabla_\theta D_{\mathrm{RKL}}^R
+
D_{\mathrm{RKL}}^R\nabla_\theta w_R^S.
\label{eq:regional_alignment_gradient}
\end{equation}
The first term ties conditional correction to the mass being
calibrated, leading to \emph{supervision masking} when student
mass is small.
For the full-vocabulary partition, regional mass gaps satisfy
\begin{equation}
w_-^S-w_-^T
=
w_+^T-w_+^S
=
0.5 \cdot \|p_T-p_S\|_1.
\label{eq:regional_weight_gap}
\end{equation}
Fig.~\ref{fig:intro_analysis}(c) further shows that conditional
discrepancies generally increase with teacher--student disagreement
while their student-mass weights decrease.
When within-region probabilities remain misaligned, even mass
calibration toward the teacher can disproportionately boost
incorrect candidates or suppress correct ones.
The second term feeds conditional discrepancies into mass updates,
creating an \emph{optimization shortcut}: reducing the alignment
penalty through lower weights without improving the conditional
distribution.
We define \emph{self-limiting supervision} as this optimization
limitation: the student's own probability allocation constrains
the correction of unresolved candidate discrepancies.
The exact equivalence in Eq.~\eqref{eq:regional_objective}
shows that this coupling is intrinsic to the original objective.

Forward KL similarly underweights negative-region alignment
with the smaller teacher mass.
JSD's mass-dependent outer weights and mixture coefficients
can weaken supervision or reduce penalties without improved alignment.
These effects reflect a shared coupling between regional probability
mass and conditional alignment across these objectives;
further analysis is provided in the \textit{supplementary material}.
\subsection{Regional Min--Max Optimization}

Supervision masking and optimization shortcuts call for reformulating
both the strength and the optimization of conditional alignment.
Selecting weights by their realized corrective effects
would require repeated trial updates and reevaluation
of conditional divergences.
Since these effects are costly to obtain before an update,
we take inspiration from prior
work~\cite{hashimoto2018fairness,sagawa2019distributionally}
and instead use the worst-case weighted conditional loss
as the optimization criterion.
Specifically, we formulate a local min--max problem over
the mass-calibration interval and possible post-update conditional
divergences, controlling this worst-case loss to strengthen
supervision over discrepancies masked by small current weights.

Let $D^R=D_{\mathrm{RKL}}^R$. The regional objective is
\begin{equation}
\min_\theta\;
\max_{\substack{
w_R\in\mathcal W_R\\
\widetilde D^R\in\mathcal U_R(\theta)
}}
w_R\widetilde D^R.
\label{eq:regional_minmax}
\end{equation}
Here, $\widetilde D^R$ denotes a potential post-update
conditional divergence, $\mathcal U_R(\theta)$ is its
uncertainty set, and $\mathcal W_R$ is the regional
mass-calibration interval.

\begin{lemma}[Local descent of the conditional divergence]
\label{lem:regional_descent}
Fix the sampled prefix, teacher distribution, and
regional partition.
Suppose $\nabla D^R$ is locally Lipschitz near $\theta$
and $\langle\nabla D^R(\theta),g\rangle>0$.
For sufficiently small $\eta>0$, the descent
lemma~\cite{nesterov2013introductory} gives
\begin{equation}
0\leq
\widetilde D^R:=D^R(\theta-\eta g)
\leq D^R(\theta).
\label{eq:regional_nonincrease}
\end{equation}
\end{lemma}

Based on this descent condition, we construct the uncertainty set
$\mathcal U_R(\theta)=[0,D^R(\theta)]$.
The current student mass $w_R^S$ and the teacher target $w_R^T$
define the starting point and target of mass calibration,
delimiting the regional calibration interval:
\begin{equation}
\mathcal W_R
=
[\min\{w_R^T,w_R^S\},\max\{w_R^T,w_R^S\}].
\label{eq:weight_calibration_range}
\end{equation}

\begin{corollary}[Closed-form maximizer]
\label{cor:regional_closed_form}
The inner maximum in Eq.~\eqref{eq:regional_minmax}
is $w_R^*D^R(\theta)$, where
\begin{equation}
w_R^*=\max\{w_R^T,w_R^S\}.
\label{eq:regional_compensation_weights}
\end{equation}
\end{corollary}

The objective therefore reduces to
\begin{equation}
\min_\theta\;
\operatorname{sg}(w_R^*)D^R(\theta),
\label{eq:regional_reduced_objective}
\end{equation}
where $\operatorname{sg}$ keeps the derived weight fixed
during each update.

This construction addresses both \emph{supervision masking}
and \emph{optimization shortcuts}.
The inner maximization evaluates within-region discrepancies
through their largest weighted conditional loss within the
calibration range, removing the masking induced by the smaller
student mass.
Fixing this weight during the update removes the direct gradient
path through the outer weight in
Eq.~\eqref{eq:regional_alignment_gradient},
blocking direct weight-reduction shortcuts.

\subsection{Final Goal}
\label{sec:final_goal}

Finally, we combine the regional min--max objective with
probability-mass calibration to obtain \textbf{A}daptive \textbf{D}ecoupled
\textbf{D}istillation (ADD).
At each generation position, ADD partitions the student's
top-$K$ candidates $\mathcal C_K$ according to
Eq.~\eqref{eq:region_partition} and aggregates the remaining
tokens into a tail bucket~\cite{yuan2026vision}.
For $a\in\{T,S\}$, define
\[
\begin{aligned}
w_{\mathrm{tail}}^a
=1-\sum_{v\in\mathcal C_K}p_a(v),\quad
\boldsymbol w^a
=(w_+^a,w_-^a,w_{\mathrm{tail}}^a).
\end{aligned}
\]
The mass-calibration term is
$\mathcal L_{\mathrm{mass}}=D_{RKL}(\boldsymbol w^S\|\boldsymbol w^T)$.
Using the adaptive weights $w_R^*$ derived above,
the final per-position objective is
\begin{equation}
\min_\theta\;
\mathcal L_{\mathrm{mass}}
+\lambda\sum_{R\in\{+,-\}}
\operatorname{sg}(w_R^*)D^R(\theta),
\label{eq:compensated_regional_objective}
\end{equation}
where $\lambda>0$ is a hyperparameter.
ADD thus mitigates \emph{self-limiting supervision} by strengthening
corrective signals for unresolved candidate discrepancies
underweighted by the student's own probability allocation
and blocking direct weight-reduction shortcuts.

\section{Experiments}
\subsection{Experimental Settings} 
\paragraph{Model Training and Implementation Details.} Our method can be readily integrated into existing visual on-policy self-distillation
frameworks. We train Qwen3.5-2B and Qwen3.5-4B~\cite{qwen3.5} for one epoch on the 6,241
synthetic visual question--answer pairs released by
Vision-OPD and evaluate the final checkpoint of each run.
The maximum prompt and response lengths are 8,192 and 1,024 tokens,
respectively. For Vision-OPD and OPD-V, we update the teacher parameters using
an exponential moving average (EMA) of the student parameters
with an update rate of $0.05$. VAD uses a frozen teacher.
We set $\lambda=0.7$ for all ADD variants.
For each comparison, we use matched training hyperparameters of the corresponding baseline. To remain consistent with the baselines, we use top-$K$ distillation with $K=100$, aggregating the remaining probability mass into a tail bucket~\cite{zaidi2025sparse}.
All experiments were conducted on a server equipped with NVIDIA RTX PRO 6000 GPUs.
Details are provided in the \textit{supplementary material}.
\paragraph{Evaluation Benchmarks.}
We evaluate fine-grained visual understanding in six settings across
four benchmark families. In V$^{*}$Bench~\cite{wu2024v}, models answer
questions about object attributes and spatial relations by locating the
relevant objects in cluttered images. ZoomBench~\cite{wei2026zooming}
tests whether models can recover evidence confined to a small part of
the image, such as text, object structure, or material cues. We use both
resolutions of HR-Bench~\cite{wang2025divide}, 4K and 8K, to assess how
well models resolve details of individual instances and combine visual
information across instances. We also evaluate on the English and
Chinese versions of MME-RealWorld~\cite{zhang2025mme}, which
cover perception and reasoning in practical domains such as scene text
recognition, chart interpretation, remote sensing, and autonomous driving. 
For evaluation, we evaluate the methods with the released Vision-OPD
pipeline, which uses a calibrated parser to recover answer
choices from generated responses. When the parser returns no valid choice,
we query DeepSeek Flash 4.1~\cite{xu2026deepseek} via its API as a fallback judge. 
\paragraph{Baselines.}
We report results for representative open-weight models and evaluate
the effect of adding our method to three visual on-policy
self-distillation methods. 
\textbf{Open-weight models.}
ZwZ-8B~\cite{wei2026zooming} distills supervision from image crops into a model that predicts from the full image.
DeepEyes-7B~\cite{zheng2026deepeyes} uses reinforcement learning to decide when and where to zoom in during reasoning.
SenseNova-MARS-8B~\cite{chng2026sensesearch} learns to coordinate image cropping with text and image search.
We also include Qwen3-VL-235B-A22B-Instruct~\cite{Qwen3-VL} and Qwen3.5-397B-A17B~\cite{qwen3.5} as large-scale multimodal reference models.
\textbf{Visual on-policy self-distillation methods.}
We use Vision-OPD, OPD-V, and VAD as baselines and evaluate
each method with and without ADD. For each
baseline, we preserve its supervision construction and
replace its final optimization objective with the
ADD objective. Consistent with these methods,
all three baselines and their ADD variants
use reverse KL as the base distillation divergence
in the main experiments.

\subsection{Experimental Results}
\begin{table*}[t]
    \centering
\caption{\textbf{Fine-grained visual understanding.}
    We report accuracy (\%) for each model. Avg. is the mean of six accuracies.
    In +ADD rows, green and orange denote gains and drops
    from the row above, with darker shades for larger changes.
    \textbf{Bold} and \underline{underlining} mark the best and second-best scores,
    respectively, within the visual self-distillation group.}
    \label{tab:main_results}
    \small
    \setlength{\tabcolsep}{4pt}
    \renewcommand{\arraystretch}{1.08}
    \definecolor{ResultGreen}{HTML}{64BC90}
    \definecolor{ResultOrange}{HTML}{E8A065}
    \newcommand{\ResultModule}{\hspace{1em}+ADD}
    \newcommand{\ResultHead}[1]{\makebox[\ResultCellWidth]{#1}}
    \newcommand{\ResultShade}[2]{%
        \begingroup
        \setlength{\fboxsep}{0pt}%
        \colorbox{#1}{\makebox[\ResultCellWidth]{\strut#2}}%
        \endgroup}
    \newcommand{\ResultUp}[2]{\ResultShade{ResultGreen!#1}{#2}}
    \newcommand{\ResultDown}[2]{\ResultShade{ResultOrange!#1}{#2}}

    \newcolumntype{M}[1]{>{\def\ResultCellWidth{#1}}c@{}}
    \begin{tabularx}{\textwidth}{@{}>{\raggedright\arraybackslash}X c @{\hspace{\tabcolsep}}
        M{50pt} M{57pt} *{4}{M{56pt}} M{38pt}}
        \toprule
        \multirow{2}{*}{Model} &
        \multirow{2}{*}{Params} &
        \ResultHead{\multirow{2}{*}{V$^*$Bench}} &
        \ResultHead{\multirow{2}{*}{ZoomBench}} &
        \multicolumn{2}{@{}c@{}}{HR-Bench} &
        \multicolumn{2}{@{}c@{}}{MME-RealWorld} &
        \ResultHead{\multirow{2}{*}{Avg.}} \\
        \cmidrule(lr){5-6}\cmidrule(lr){7-8}
        & & & & 4K & 8K & EN & CN & \\
        \midrule
        \multicolumn{9}{@{}c@{}}{Open-weight baselines} \\
        \midrule[0.35pt]
        DeepEyes          & 7B   & 85.56 & 46.51 & 75.13 & 72.63 & 64.10 & 64.09 & 68.00 \\
        SenseNova-MARS    & 8B   & 92.15 & 47.81 & 83.13 & 78.38 & 67.90 & 68.90 & 73.05 \\
        ZwZ               & 8B   & 87.96 & 56.69 & 83.63 & 81.75 & 66.57 & 68.09 & 74.12 \\
        Qwen3-VL-Instruct & 235B & 87.37 & 57.23 & 88.00 & 79.25 & 71.09 & 65.86 & 74.80 \\
        Qwen3.5           & 397B & 87.96 & 57.16 & 89.38 & 85.50 & 74.82 & 69.82 & 77.44 \\
        \midrule
        \multicolumn{9}{@{}c@{}}{\textbf{Visual On-Policy Self-Distillation on Qwen3.5}} \\
        \midrule[0.35pt]
        Qwen3.5           & 2B   & 76.44 & 44.62 & 76.75 & 76.25 & 67.55 & 53.08 & 65.78 \\
        \addlinespace[1pt]
        Vision-OPD        & 2B   & 84.82 & 50.41 & 73.88 & 73.00 & 69.95 & 59.10 & 68.53 \\
        \ResultModule & 2B &
            \ResultUp{33}{87.43} & \ResultUp{18}{50.65} & \ResultUp{22}{74.63} & \ResultUp{21}{73.63} &
            \ResultDown{18}{69.80} & \ResultUp{23}{60.16} & \ResultUp{22}{69.38} \\
        \addlinespace[1pt]
        OPD-V             & 2B   & 83.77 & 53.73 & 76.63 & 76.50 & 71.63 & 60.47 & 70.46 \\
        \ResultModule & 2B &
            \ResultUp{45}{88.48} & \ResultDown{21}{53.14} & \ResultUp{19}{77.00} & \ResultUp{22}{77.38} &
            \ResultUp{25}{72.92} & \ResultUp{24}{61.67} & \ResultUp{25}{71.77} \\
        \addlinespace[1pt]
        VAD               & 2B   & 86.39 & 54.08 & 80.50 & 76.63 & 71.57 & 61.01 & 71.70 \\
        \ResultModule & 2B &
            \ResultUp{30}{88.48} & \ResultUp{19}{54.44} & \ResultUp{22}{81.25} & \ResultUp{27}{78.25} &
            \ResultUp{30}{73.80} & \ResultUp{25}{62.33} & \ResultUp{25}{73.09} \\
        \midrule[0.35pt]
        Qwen3.5           & 4B   & 82.20 & 48.79 & 86.12 & 79.62 & 58.81 & 60.44 & 69.33 \\
        \addlinespace[1pt]
        Vision-OPD        & 4B   & 89.53 & 59.88 & 83.00 & 80.13 & 75.38 & 70.49 & 76.40 \\
        \ResultModule & 4B &
            \ResultUp{36}{\underline{92.67}} & \ResultUp{22}{\underline{60.71}} & \ResultDown{21}{82.38} & \ResultUp{28}{81.88} &
            \ResultUp{19}{75.79} & \ResultUp{17}{70.53} & \ResultUp{23}{77.33} \\
        \addlinespace[1pt]
        OPD-V             & 4B   & 91.10 & 59.29 & 82.75 & 80.50 & 76.17 & 72.15 & 76.99 \\
        \ResultModule & 4B &
            \ResultUp{23}{92.15} & \ResultUp{23}{60.36} & \ResultUp{28}{84.63} & \ResultUp{22}{81.25} &
            \ResultUp{20}{\underline{76.67}} & \ResultUp{20}{72.65} & \ResultUp{23}{77.95} \\
        \addlinespace[1pt]
        VAD               & 4B   & 92.15 & 59.41 & \underline{86.88} & \textbf{84.00} & 76.23 & \underline{72.93} & \underline{78.60} \\
        \ResultModule & 4B &
            \ResultUp{26}{\textbf{93.71}} & \ResultUp{27}{\textbf{61.07}} & \ResultUp{22}{\textbf{87.75}} & \ResultDown{21}{\underline{83.38}} &
            \ResultUp{21}{\textbf{76.86}} & \ResultUp{23}{\textbf{73.89}} & \ResultUp{22}{\textbf{79.44}} \\
        \bottomrule
    \end{tabularx}
\end{table*}

Table~\ref{tab:main_results} compares Vision-OPD, OPD-V, and VAD with and without ADD on fine-grained visual
understanding. Building on each method's original design, ADD improves average accuracy for all three methods at both model sizes. In the same method order, the gains are 0.85, 1.31, and 1.39 percentage points on Qwen3.5-2B, and 0.93, 0.96, and 0.84 points on Qwen3.5-4B.

ADD improves V$^*$Bench accuracy in all six paired comparisons, with the largest gain of 4.71 points for OPD-V-2B (83.77\% to 88.48\%). VAD-4B with ADD achieves the highest overall mean accuracy (79.44\%) and leads on V$^*$Bench, ZoomBench, and both language versions of MME-RealWorld. The gains across all three methods at both model sizes suggest that ADD complements existing designs
for visual on-policy self-distillation effectively.

\subsection{Ablation Study}

All ablations use Qwen3.5-4B under the Vision-OPD setup.

\begin{figure}
    \centering
    \includegraphics[width=1.0\linewidth]{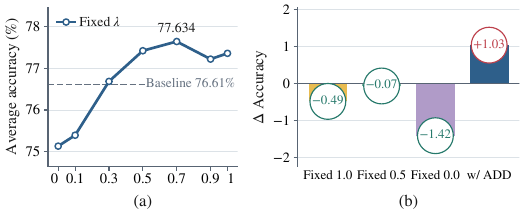}
    \caption{
        (a) Mean accuracy versus $\lambda$.
        (b) Mean accuracy changes from Vision-OPD 
        for fixed regional weights and ADD.
        Both panels average accuracy over five settings:
        V$^*$Bench, ZoomBench, HR-Bench 4K/8K, and MME-RealWorld CN.
    }
    \label{fig:weight_ablation}
\end{figure}

\paragraph{Component ablation.}
Table~\ref{tab:sg_ablation_4b} evaluates adaptive weights $w_R^*$
and stop-gradient (sg) by varying the regional weighting rule
and the gradient path through its weights.
With sg enabled, $w_R^*$ improves all four metrics over the original
regional weights, with a 2.25-point gain on ZoomBench.
Adaptive weighting improves performance beyond gradient
decoupling alone, demonstrating the value of adjusting conditional
supervision strength.
With $w_R^*$ enabled, sg improves V$^*$Bench, ZoomBench,
and HR-Bench Avg.\ by 1.57, 0.48, and 0.94 percentage points,
respectively.
Weights are recomputed from current predictions at every
step; sg removes only the gradient path through the outer weights
during backpropagation.
It therefore preserves dynamic weighting while blocking direct
weight-reduction shortcuts.
The two components address supervision strength and the optimization
path, respectively, working together to improve the correction
of unresolved candidate discrepancies. 
\begin{table}[t]
    \centering
    \caption{
        Component ablation of adaptive regional weights $w_R^*$
        and stop-gradient (sg).
        $\checkmark$ and $\times$ indicate enabled and disabled components.
        We report accuracy (\%); HR-Bench Avg.\ averages the 4K and 8K accuracies.
    }
    \label{tab:sg_ablation_4b}
    \small
    \setlength{\tabcolsep}{2pt}
    \renewcommand{\arraystretch}{1.12}
    \begin{tabular*}{\columnwidth}{@{}cc@{\extracolsep{\fill}}cccc@{}}
        \toprule
        \multirow{2}{*}{$w_R^*$} &
        \multirow{2}{*}{sg} &
        \multirow{2}{*}{V$^*$} &
        \multirow{2}{*}{ZoomBench} &
        HR-Bench & MME-RealWorld \\
        & & & & Avg. & CN \\
        \midrule
        $\times$ & $\times$ &
        89.53 & 59.88 & 81.57 & 70.49 \\
        $\times$ & $\checkmark$ &
        91.10 & 58.46 & 81.63 & 70.22 \\
        $\checkmark$ & $\times$ &
        91.10 & 60.23 & 81.19 & \textbf{70.63} \\
        $\checkmark$ & $\checkmark$ &
        \textbf{92.67} & \textbf{60.71} &
        \textbf{82.13} & 70.53 \\
        \bottomrule
    \end{tabular*}
\end{table}
\vspace{-0.7pt}
\paragraph{Importance of adaptive weighting.}
Supervision masking varies across positions and is particularly
pronounced at the few positions carrying most useful visual
supervision, calling for per-position adaptation.
In Fig.~\ref{fig:weight_ablation}(b), all three fixed configurations ($w_+=w_-=c$)
underperform Vision-OPD, whereas ADD improves mean accuracy by
1.03 percentage points across five settings:
V$^*$Bench, ZoomBench, HR-Bench 4K, HR-Bench 8K,
and MME-RealWorld CN.
We further replace each region's weights with their mean over valid
positions within each sequence, preserving mean weights and their
adaptation during training while removing positional variation.
Per-position weights improve all metrics, including gains
of 5.24 and 3.67 percentage points on V$^*$Bench and ZoomBench,
respectively (Table~\ref{tab:sequence_weight_ablation_4b}).
The results demonstrate the benefit of adapting correction
strength to the uneven supervision masking across positions.

\begin{table}[t]
    \centering
    \caption{
        Ablation of the granularity of regional weight adaptation.
        Sequence means retain weight changes during training
        but remove variation within each sequence.
        We report accuracy (\%); HR-Bench Avg.\ averages the 4K and 8K accuracies.
    }
    \label{tab:sequence_weight_ablation_4b}
    \footnotesize
    \setlength{\tabcolsep}{1.5pt}
    \renewcommand{\arraystretch}{1.12}
    \begin{tabular*}{\columnwidth}{@{}l@{\extracolsep{\fill}}cccc@{}}
        \toprule
        \multirow{2}{*}{Weight assignment} &
        \multirow{2}{*}{V$^*$} &
        \multirow{2}{*}{ZoomBench} &
        HR-Bench & MME-RealWorld \\
        & & & Avg. & CN \\
        \midrule
        Sequence mean &
        87.43 & 57.04 & 80.82 & 70.49 \\
        ADD &
        \textbf{92.67} & \textbf{60.71} &
        \textbf{82.13} & \textbf{70.53} \\
        \bottomrule
    \end{tabular*}
\end{table}
\vspace{-0.3pt}
\paragraph{Hyperparameter $\lambda$.}
Fig.~\ref{fig:weight_ablation}(a) examines the conditional-alignment
coefficient $\lambda$.
Setting $\lambda=0$ retains only mass calibration and underperforms
Vision-OPD, showing that conditional alignment is an important
component of ADD's gains.
ADD outperforms Vision-OPD across the tested values from 0.3 to 1.0,
demonstrating that its gains persist over a broad range
of alignment strengths. Additional experiments are provided
in the \textit{supplementary material}.
\section{Discussion}

\begin{table}[t]
    \centering
    \caption{Comparison under different divergence objectives.}
    \label{tab:divergence_comparison}
    \small
    \setlength{\tabcolsep}{3pt}
    \renewcommand{\arraystretch}{1.12}
    \begin{tabular*}{\linewidth}{
        @{\extracolsep{\fill}}l *{6}{c}@{}
    }
        \toprule
        \multirow{2}{*}{\textbf{Method}} &
        \multicolumn{3}{c}{\textbf{V$^*$Bench}} &
        \multicolumn{3}{c}{\textbf{ZoomBench}} \\
        \cmidrule(lr){2-4}
        \cmidrule(lr){5-7}
        & JSD & FKL & RKL & JSD & FKL & RKL \\
        \midrule
        Vision-OPD   & 90.58 & 88.48 & 89.53 & 59.17 &\textbf{58.93} & 59.88 \\
        $\quad$+ADD & \textbf{91.62} & \textbf{90.58} & \textbf{92.67} & \textbf{59.29} & 58.82 & \textbf{60.71} \\
        \bottomrule
    \end{tabular*}
\end{table}

\begin{table*}[t]
    \centering
    
    \caption{Evaluation results using \textbf{Qwen3-4B-Instruct-2507} as the teacher~\cite{yang2025qwen3}.
Mathematical reasoning results are reported as
\textbf{mean@16} accuracy (\%), while code generation results
are reported as \textbf{mean@4} accuracy (\%).
Math Avg and Code Avg denote the arithmetic mean over the corresponding
benchmarks. \textbf{Bold} values indicate the best result within each
student group.}
    \small
    \setlength{\tabcolsep}{5pt}
    \renewcommand{\arraystretch}{1.12}
    \begin{tabular*}{\textwidth}{
        @{\extracolsep{\fill}}l l *{7}{c}@{}
    }
        \toprule
        \multirow{2}{*}{\textbf{Student}} &
        \multirow{2}{*}{\textbf{Method}} &
        \multicolumn{4}{c}{\textbf{Math} (mean@16)} &
        \multicolumn{3}{c}{\textbf{Code} (mean@4)} \\
        \cmidrule(lr){3-6}
        \cmidrule(lr){7-9}
        & & AIME24 & AIME25 & HMMT25 & Math Avg
          & HE+ & MBPP+ & Code Avg \\
        \midrule

        \multicolumn{2}{l}{\textit{Teacher Model}} &
        61.45 & 45.00 & 29.17 & 45.21 &
        79.57 & 73.61 & 76.59 \\
        \midrule

        \multirow{3}{*}{Qwen3-1.7B}
        & Base
        & 11.88 & 10.00 & 3.54 & 8.47
        & 59.60 & 53.11 & 56.36 \\
        & OPD
        & 33.54 & 20.62 & 12.71 & 22.29
        & 61.28 & 52.65 & 56.97 \\
        & \textbf{$\quad$+ADD}
        & \textbf{35.00} & \textbf{21.04}
        & \textbf{13.96} & \textbf{23.33}
        & \textbf{64.18} & \textbf{53.84}
        & \textbf{59.01} \\
        \midrule

        \multirow{3}{*}{Qwen3-0.6B}
        & Base
        & 1.25 & 1.46 & 0.42 & 1.04
        & 28.51 & 27.38 & 27.95 \\
        & OPD
        & 8.75 & 15.21 & 8.33 & 10.76
        & 38.11 & 31.15 & 34.63 \\
        & \textbf{$\quad$+ADD}
        & \textbf{10.62} & \textbf{17.29}
        & \textbf{8.54} & \textbf{12.15}
        & \textbf{38.57} & \textbf{33.73}
        & \textbf{36.15} \\
        \bottomrule
    \end{tabular*}
    \label{tab:extend_exp}
\end{table*}
\subsection{Effectiveness of ADD}

\begin{figure*}[t]
    \centering
    \includegraphics[width=\textwidth]{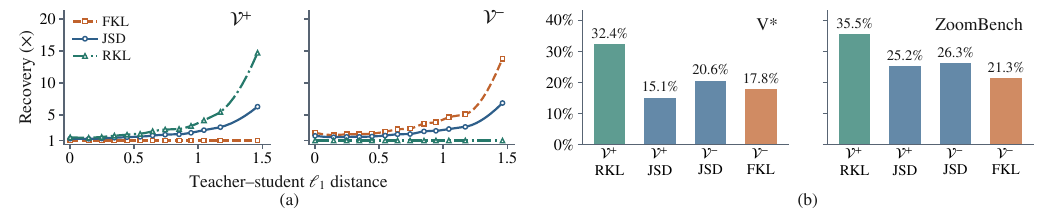}
\caption{
(a) Ratios of summed regional conditional losses
(ADD/baseline, excluding $\lambda$) across teacher--student
$\ell_1$ bins.
(b) Conditional divergence reductions (\%) with ADD
relative to Vision-OPD on V$^*$Bench and ZoomBench.
}
    \label{fig:regional-loss-recovery}
\end{figure*}

We examine how ADD mitigates \emph{self-limiting supervision}
by relating supervision recovery to conditional alignment.
Fig.~\ref{fig:regional-loss-recovery}(a) reports the ratio
of summed regional conditional losses under ADD to those
under the baseline within each teacher--student $\ell_1$
distance bin, excluding the global coefficient $\lambda$.
For underweighted alignment terms, the ratio increases
with disagreement, indicating greater relative reinforcement
at high-disagreement positions.
FKL in $\mathcal{V}^{+}$ and RKL in $\mathcal{V}^{-}$
remain at $1\times$, since their baseline weights
already equal those of ADD.

To examine whether stronger supervision translates into
better conditional alignment, we select V*Bench and
ZoomBench, whose target crop regions enable the construction
of a privileged teacher.
We fix the candidate sets and regional partitions using
the teacher and initial student predictions, providing
a common basis for comparing the trained models.
In this diagnostic, ADD reduces positive-region RKL
by 32.4\% and 35.5\% on the two benchmarks, respectively
(Fig.~\ref{fig:regional-loss-recovery}(b)).
Because these conditional divergences exclude outer
regional weights, their reductions show that supervision
recovery accompanies closer alignment of relative candidate
probabilities, providing evidence that ADD mitigates
self-limiting supervision.

As discussed in Sec.~\ref{coupledalign}, FKL underweights
negative-region alignment with the smaller teacher mass,
while JSD's mass-dependent outer weights and mixture coefficients
can weaken supervision or reduce penalties without improving
conditional alignment.
We extend ADD to both objectives to address these forms of
\emph{self-limiting supervision}.
Table~\ref{tab:divergence_comparison} shows that ADD improves
V$^*$Bench accuracy under all three objectives and ZoomBench
accuracy under JSD and RKL, with gains in five of six comparisons.
Fig.~\ref{fig:regional-loss-recovery}(b) further shows lower
negative-region FKL and lower JSD in both regions on both
benchmarks, supporting improved within-region alignment.

These accuracy comparisons and alignment diagnostics
provide evidence for ADD's applicability across divergence
objectives and its correction of conditional distributions,
respectively. Further analyses and experimental results are provided in the \textit{supplementary material}.

\subsection{Attention Visualization}
\begin{figure}[!tb]
    \centering
    \includegraphics[width=\linewidth]{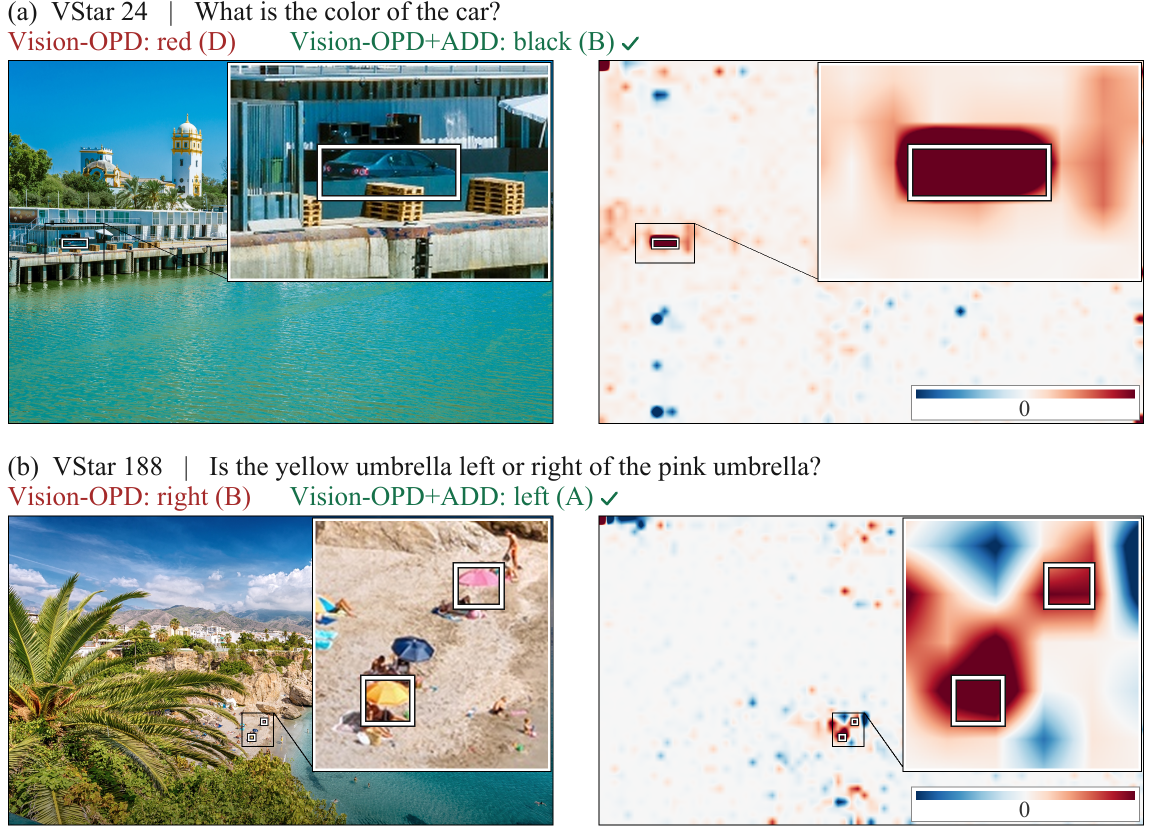}
    \caption{Case Study on V$^*$Bench. Response-averaged image attention differences from independent
    generations (Vision-OPD+ADD minus Vision-OPD).
    Red/blue indicate higher/lower attention with ADD.}
    \label{fig:attention_visualization}
\end{figure}
To analyze the source of performance gains,
we examine how the trained models use visual evidence.
Vision-OPD and Vision-OPD+ADD independently generate responses to the same image and question. Given Qwen3.5’s architecture~\cite{qwen3.5}, we average each generated token’s attention to image tokens across the eight observable full-attention layers and 16 query heads~\cite{jiang2025tarac,yu2024attention}. We then average over all generated tokens using each model's own response length. Fig.~\ref{fig:attention_visualization} shows the attention difference on the native visual grid, computed as Vision-OPD+ADD minus Vision-OPD. Red and blue indicate higher attention with and without ADD, respectively.
In Fig.~\ref{fig:attention_visualization}(a), ADD assigns higher mean attention to the small car and correctly identifies its color as black, whereas Vision-OPD predicts red. In (b), increased attention over both umbrellas accompanies the correct judgment that the yellow umbrella is to the left of the pink one, correcting Vision-OPD's reversed relation. These examples are consistent with improved visual grounding for both attribute recognition and spatial reasoning.

\subsection{Extension to More Tasks}

Although ADD was motivated by fine-grained visual
self-distillation, the coupling between probability mass
and conditional alignment can also cause
\emph{self-limiting supervision} in non-visual on-policy
distillation.
ADD's regional decomposition and alignment weights depend only
on teacher--student predictive distributions, making it directly
applicable to these tasks.
Following the experimental setup
of~\cite{agarwal2024policy,yang2026learning}, we evaluate
mathematical reasoning (AIME24, AIME25, and
HMMT25~\cite{balunovic2026matharena}) and code generation
(HE+ and MBPP+~\cite{liu2023your}), reporting mean@16 and
mean@4 accuracy, respectively.
Mean@$K$ denotes the average accuracy over $K$ independently
sampled responses per problem.
As shown in Table~\ref{tab:extend_exp}, ADD outperforms
standard OPD (GKD$_{\mathrm{RKL}}$~\cite{agarwal2024policy})
on all five benchmarks at both student model sizes.
These consistent gains support ADD's applicability across tasks
and suggest that \emph{self-limiting supervision} may be a shared
optimization issue in on-policy distillation.

\section{Conclusion}

We identified \emph{self-limiting supervision} in visual on-policy
self-distillation, where the student's own probability allocation
can constrain the correction of unresolved candidate discrepancies.
An exact regional decomposition reveals the coupling between
probability-mass calibration and conditional alignment, which can
weaken supervision and allow alignment penalties to decrease through
weight reduction without correcting the discrepancies. We proposed
ADD, which derives adaptive regional weights through min--max
optimization and holds them fixed during each update, decoupling
conditional alignment from mass calibration without additional
forward passes. Experiments show that ADD improves average accuracy
across six fine-grained visual evaluation settings for Vision-OPD,
OPD-V, and VAD across different model sizes, with further gains in
more tasks. Conditional-divergence
diagnostics further show improved alignment of relative candidate
probabilities.

\section*{LLM Usage Statement}
OpenAI Codex GPT-6.1 Sol High was used for English language editing.
DeepSeek V4.1 Flash served as the evaluation judge, replacing
the locally deployed \texttt{openai/gpt-oss-120b} used in~\cite{yuan2026vision}.
{
    \small
    \bibliographystyle{ieeenat_fullname}
    \bibliography{main}
}
\maketitlesupplementary
\appendix

\section{Self-Limiting Supervision}
\label{sec:supp_regional_decomposition}

At a fixed student-generated prefix, the crop-conditioned teacher
$p_T$ and full-image student $p_S$ can disagree on how much
probability to give a set of candidates and how to distribute that
probability among them. Under RKL, the student weights the second
error by the mass it assigns to the set. To see how this weighting
works for each divergence, we split the vocabulary by the sign
of the probability gap,
\begin{equation}
\begin{aligned}
 \mathcal V^+ &= \{v\in\mathcal V : p_T(v)\geq p_S(v)\},\\
 \mathcal V^- &= \{v\in\mathcal V : p_T(v)<p_S(v)\}.
\end{aligned}
\label{eq:supp_partition}
\end{equation}
For $a\in\{T,S\}$ and $R\in\{+,-\}$, define
\begin{equation}
 w_R^a=\sum_{v\in\mathcal V^R}p_a(v),
 \qquad p_a^R(v)=\frac{p_a(v)}{w_R^a}
 \quad (v\in\mathcal V^R).
\label{eq:supp_factorization}
\end{equation}
For $v\in\mathcal V^R$, $p_a(v)=w_R^a p_a^R(v)$ and
$\sum_{u\in\mathcal V^R}p_a^R(u)=1$. Write $\boldsymbol w^a=(w_+^a,w_-^a)$.
Ties belong to $\mathcal V^+$. Since the probability gaps sum to zero, the
student's mass deficit in $\mathcal V^+$ equals its excess in
$\mathcal V^-$,
\begin{equation}
 w_+^T-w_+^S = w_-^S-w_-^T = \tfrac12\|p_T-p_S\|_1 .
\label{eq:supp_mass_gap}
\end{equation}
This identity uses the full vocabulary. The gradient calculations
below hold the prefix, teacher distribution, and regional partition
fixed.

\subsection{Reverse KL}
\label{sec:supp_rkl}

\paragraph{Regional decomposition.}
Reverse KL takes the expectation of $\log(p_S/p_T)$ under the
student. For $v\in\mathcal V^R$, substitute $p_S(v)=w_R^S p_S^R(v)$
and $p_T(v)=w_R^T p_T^R(v)$ in the contribution of region $R$.
Splitting the log ratio then gives
\begin{equation}
\begin{aligned}
 &\sum_{v\in\mathcal V^R} p_S(v)\log\frac{p_S(v)}{p_T(v)}\\
 &=w_R^S\sum_{v\in\mathcal V^R}p_S^R(v)
   \left[\log\frac{w_R^S}{w_R^T}+\log\frac{p_S^R(v)}{p_T^R(v)}\right]\\
 &=w_R^S\log\frac{w_R^S}{w_R^T}
   +w_R^S\,\mathrm{KL}(p_S^R\|p_T^R).
\end{aligned}
\label{eq:supp_rkl_one_region}
\end{equation}
The first log ratio is constant within $R$ and $p_S^R$ sums to one.
The second log ratio compares relative probabilities within $R$.
Writing $D_{\mathrm{RKL}}=\mathrm{KL}(p_S\|p_T)$ and
$D_{\mathrm{RKL}}^R=\mathrm{KL}(p_S^R\|p_T^R)$, summation over
the two regions yields
\begin{equation}
 D_{\mathrm{RKL}}
 =\mathrm{KL}(\boldsymbol w^S\|\boldsymbol w^T)
 +\sum_R w_R^S D_{\mathrm{RKL}}^R .
\label{eq:supp_rkl_decomposition}
\end{equation}
The first term calibrates $w_R^S$ toward $w_R^T$. The second uses
the same $w_R^S$ to weight the conditional error. Differentiating
this product gives the two terms behind self-limiting supervision,
\begin{equation}
\begin{aligned}
 \nabla_\theta(w_R^S D_{\mathrm{RKL}}^R)
 &=w_R^S\nabla_\theta D_{\mathrm{RKL}}^R
   +D_{\mathrm{RKL}}^R\nabla_\theta w_R^S .
\end{aligned}
\label{eq:supp_rkl_gradient}
\end{equation}

\paragraph{Supervision masking.}
In Eq.~\eqref{eq:supp_rkl_gradient}, $w_R^S$ scales the gradient
that corrects relative probabilities. Fig.~1(c) of the main paper
shows the consequence for teacher-favored candidates. At positions
with greater disagreement, the student's relative probabilities are
typically further from the teacher's (larger $D_{\mathrm{RKL}}^+$),
yet the student gives their correction less weight (smaller $w_+^S$).

\paragraph{Optimization shortcut.}
The student sets the weight on its own regional error through
$w_R^S$, giving the second term in Eq.~\eqref{eq:supp_rkl_gradient}.
Suppose it halves the probability of every candidate in $\mathcal V^+$
and redistributes the freed mass within $\mathcal V^-$. The relative
probabilities in $\mathcal V^+$ stay the same. The error
$D_{\mathrm{RKL}}^+$ therefore stays the same, but its contribution
$w_+^S D_{\mathrm{RKL}}^+$ falls by half.

At a prefix where teacher and student disagree, transfer
$0<\varepsilon<w_+^S$ from the positive region to the negative
region, keeping $p_S^+$ and $p_S^-$ fixed.
The two student masses become $w_+^S-\varepsilon$ and
$w_-^S+\varepsilon$. By Eq.~\eqref{eq:supp_rkl_decomposition}, the
change in the sum of the conditional penalties is

\begin{equation}
 \Delta\!\left(\sum_R w_R^S D_{\mathrm{RKL}}^R\right)
 =\varepsilon\bigl(D_{\mathrm{RKL}}^--D_{\mathrm{RKL}}^+\bigr).
\label{eq:supp_rkl_conditional_transfer}
\end{equation}

If $D_{\mathrm{RKL}}^+>D_{\mathrm{RKL}}^-$, this sum falls. The
mass-calibration term also changes, since the student has taken
probability away from candidates it already underestimates. Let
$\ell(v)=\log[p_S(v)/p_T(v)]$ before the transfer. Differentiating
the mass term and adding the slope of the conditional sum gives

\begin{equation}
\begin{aligned}
 \left.\frac{\mathrm dD_{\mathrm{RKL}}}{\mathrm d\varepsilon}\right|_{0}
 &=\log\frac{w_-^S w_+^T}{w_-^T w_+^S}
   +D_{\mathrm{RKL}}^--D_{\mathrm{RKL}}^+\\
 &=\mathbb E_{p_S^-}[\ell]-\mathbb E_{p_S^+}[\ell]>0 .
\end{aligned}
\label{eq:supp_rkl_transfer}
\end{equation}

The second line uses $\mathbb E_{p_S^R}[\ell]
=\log(w_R^S/w_R^T)+D_{\mathrm{RKL}}^R$. Every candidate receiving
mass has $\ell(v)>0$, and every candidate giving up mass has
$\ell(v)\leq0$. The full RKL therefore increases along this
direction, even when the weighted conditional sum falls.

\subsection{Forward KL}
\label{sec:supp_fkl}

\paragraph{Separating mass and conditional probabilities.}
Let $D_{\mathrm{FKL}}=\mathrm{KL}(p_T\|p_S)$. For each token in
region $R$, Eq.~\eqref{eq:supp_factorization} gives
\begin{equation}
 \log\frac{p_T(v)}{p_S(v)}
 =\log\frac{w_R^T}{w_R^S}+\log\frac{p_T^R(v)}{p_S^R(v)}.
\label{eq:supp_fkl_log_ratio}
\end{equation}
The first log ratio is constant across candidates in $R$. Summing
over the region and using the normalization of $p_T^R$ gives
\begin{equation}
\begin{aligned}
 &\sum_{v\in\mathcal V^R}p_T(v)\log\frac{p_T(v)}{p_S(v)}\\
 &=w_R^T\sum_{v\in\mathcal V^R}p_T^R(v)
   \left[\log\frac{w_R^T}{w_R^S}+\log\frac{p_T^R(v)}{p_S^R(v)}\right]\\
 &=w_R^T\log\frac{w_R^T}{w_R^S}
   +w_R^T\,\mathrm{KL}(p_T^R\|p_S^R).
\end{aligned}
\label{eq:supp_fkl_one_region}
\end{equation}
Summing the first term over $R$ recovers the KL between the two
regional mass vectors. Writing $D_{\mathrm{FKL}}^R=
\mathrm{KL}(p_T^R\|p_S^R)$, we obtain
\begin{equation}
 D_{\mathrm{FKL}} =
 \mathrm{KL}(\boldsymbol w^T\|\boldsymbol w^S)
 +\sum_R w_R^T D_{\mathrm{FKL}}^R .
\label{eq:supp_fkl_decomposition}
\end{equation}

\paragraph{Supervision masking in the negative region.}
The student assigns excess probability to the candidates in
$\mathcal V^-$. FKL weights the error in their relative probabilities
by the teacher's mass $w_-^T$. Let $z$ denote the student logits.
Since $p_S^R$ is the softmax over logits in $\mathcal V^R$, the
conditional gradient for $v\in\mathcal V^R$ is
\begin{equation}
 \frac{\partial (w_R^T D_{\mathrm{FKL}}^R)}{\partial z_v}
 =w_R^T\bigl(p_S^R(v)-p_T^R(v)\bigr).
\label{eq:supp_fkl_gradient}
\end{equation}
The mass term contributes $(w_R^S-w_R^T)p_S^R(v)$ to the same
logit gradient. This is positive for every $v\in\mathcal V^-$, so
mass calibration pushes down the logits of the overestimated
candidates. How the student distributes probability among them is
supervised by Eq.~\eqref{eq:supp_fkl_gradient}. When $w_-^T$ is
small, the student receives little correction on their relative
probabilities, even if these candidates account for a large share
of its predictions.

\subsection{Jensen--Shannon Divergence}
\label{sec:supp_jsd}

\paragraph{Conditioning the mixture.}
Symmetric JSD compares each distribution to the same mixture
$m=(p_T+p_S)/2$,
\begin{equation}
 D_{\mathrm{JSD}}
 =\tfrac12\mathrm{KL}(p_T\|m)+\tfrac12\mathrm{KL}(p_S\|m).
\label{eq:supp_js_definition}
\end{equation}
Summing $m(v)$ over $\mathcal V^R$ gives
$w_R^M=(w_R^T+w_R^S)/2$. Dividing $m(v)$ by this mass gives
\begin{equation}
\begin{aligned}
 m^R(v)&=\frac{w_R^T p_T^R(v)+w_R^S p_S^R(v)}{w_R^T+w_R^S}\\
       &=\alpha_R p_T^R(v)+(1-\alpha_R)p_S^R(v),\\
 \alpha_R&=\frac{w_R^T}{w_R^T+w_R^S}.
\end{aligned}
\label{eq:supp_js_region_mix}
\end{equation}
The full-vocabulary mixture gives each model half the weight.
Within a region where $w_R^T=3w_R^S$, for example, the teacher
supplies three quarters of the reference distribution $m^R$.

\paragraph{Decomposing the two KL terms.}
Apply Eq.~\eqref{eq:supp_fkl_one_region} to $(p_T,m)$ and $(p_S,m)$.
With $\boldsymbol w^M=(\boldsymbol w^T+\boldsymbol w^S)/2$, each
$a\in\{T,S\}$ gives
\begin{equation}
\begin{aligned}
 \mathrm{KL}(p_a\|m)
 &=\mathrm{KL}(\boldsymbol w^a\|\boldsymbol w^M)\\
 &\quad+\sum_R w_R^a\,\mathrm{KL}(p_a^R\|m^R).
\end{aligned}
\label{eq:supp_js_two_kl}
\end{equation}
Taking their half-sum in Eq.~\eqref{eq:supp_js_definition} yields
$\mathrm{JS}_{1/2}(\boldsymbol w^T,\boldsymbol w^S)$ for the mass
term. The conditional contribution of region $R$ is
\begin{equation}
\begin{aligned}
 C_R&=\tfrac{w_R^T}{2}\,\mathrm{KL}(p_T^R\|m^R)
      +\tfrac{w_R^S}{2}\,\mathrm{KL}(p_S^R\|m^R)\\
    &=w_R^M\,\mathrm{JS}_{\alpha_R}(p_T^R,p_S^R).
\end{aligned}
\label{eq:supp_js_conditional}
\end{equation}
Here $\mathrm{JS}_{\alpha}(P,Q)$ weights the two KL terms to
$\alpha P+(1-\alpha)Q$ by $\alpha$ and $1-\alpha$. The second line
uses $w_R^M\alpha_R=w_R^T/2$ and $w_R^M(1-\alpha_R)=w_R^S/2$.
Combining mass and conditional terms gives the exact decomposition
\begin{equation}
 D_{\mathrm{JSD}}
 =\mathrm{JS}_{1/2}(\boldsymbol w^T,\boldsymbol w^S)
  +\sum_R w_R^M\,\mathrm{JS}_{\alpha_R}(p_T^R,p_S^R).
\label{eq:supp_js_decomposition}
\end{equation}

\paragraph{Supervision masking under mass imbalance.}
In Eq.~\eqref{eq:supp_js_decomposition}, a small $w_R^M$ downweights
the entire conditional penalty. An imbalance between $w_R^T$ and
$w_R^S$ also changes the divergence being weighted. The reference
$m^R$ favors the teacher in $\mathcal V^+$ ($\alpha_+\geq 1/2$)
and the student in $\mathcal V^-$ ($\alpha_-\leq 1/2$).
For $0<\alpha<1$, the mixture $\alpha P+(1-\alpha)Q$ dominates
$\alpha P$ and $(1-\alpha)Q$ pointwise. Bounding the two log ratios
by $-\log\alpha$ and $-\log(1-\alpha)$ gives
\begin{equation}
 \mathrm{JS}_\alpha(P,Q)
 \leq -\alpha\log\alpha-(1-\alpha)\log(1-\alpha).
\label{eq:supp_js_entropy_bound}
\end{equation}
When one model supplies almost all the regional mass, the reference
mixture approaches that model's conditional distribution and the
other model receives little weight. As $\alpha_R$ approaches $0$
or $1$, the bound tends to zero for any fixed pair of conditionals.
A large imbalance in mass can thus hide a persistent disagreement
about the candidates in $R$.

\paragraph{Optimization shortcut through regional mass.}
Reducing $w_R^S$ changes both the outer weight $w_R^M$ and the
reference mixture $m^R$. To find their combined effect on $C_R$,
hold $p_T^R,p_S^R,w_R^T$ fixed. In Eq.~\eqref{eq:supp_js_conditional},
the derivative of the explicit weight contributes
$\tfrac12\mathrm{KL}(p_S^R\|m^R)$. The derivatives through $m^R$
sum to
\begin{equation}
\begin{aligned}
 &-\sum_{v\in\mathcal V^R}
 \frac{w_R^T p_T^R(v)+w_R^S p_S^R(v)}{2m^R(v)}
 \frac{\partial m^R(v)}{\partial w_R^S}\\
 &\qquad=-w_R^M\frac{\partial}{\partial w_R^S}
     \sum_{v\in\mathcal V^R}m^R(v)=0 .
\end{aligned}
\label{eq:supp_js_cancellation}
\end{equation}
The last equality uses $\sum_v m^R(v)=1$ for every $w_R^S$.
The remaining derivative is
\begin{equation}
 \frac{\partial C_R}{\partial w_R^S}
 =\tfrac12\,\mathrm{KL}(p_S^R\|m^R)\geq 0 .
\label{eq:supp_js_mass_derivative}
\end{equation}
When the teacher and student disagree within $R$, this derivative
is positive. The student can therefore lower the regional JSD
penalty simply by assigning less probability to $R$, even as its
relative probabilities within $R$ stay exactly the same.

\paragraph{ADD across divergence objectives.}
RKL, FKL, and JSD each contain a term that calibrates regional
masses and a term that aligns relative probabilities within each
region. ADD sets the outer weight on each conditional divergence to
$w_R^*=\max\{w_R^T,w_R^S\}$ using the masses alone.
Under RKL this raises the positive-region
weight from $w_+^S$ to $w_+^T$. Under FKL it raises the negative-region
weight from $w_-^T$ to $w_-^S$. For JSD, ADD uses the equal-weight
$\mathrm{JS}_{1/2}(p_T^R,p_S^R)$ within each region and increases
the outer weight by $w_R^*-w_R^M=\tfrac12|w_R^T-w_R^S|$.
During backpropagation
ADD holds $w_R^*$ fixed, closing the gradient path through the
outer weight while the conditional divergence trains the student's
relative probabilities.

In the main paper's Discussion, Fig.~4(a) shows that the ratio of
ADD to baseline conditional loss grows with teacher--student
disagreement for the underweighted terms. On both V$^*$Bench and
ZoomBench, ADD reduces conditional RKL in $\mathcal V^+$, conditional
FKL in $\mathcal V^-$, and conditional JSD in both regions relative
to Vision-OPD (Fig.~4(b) of the main paper). Accuracy improves in five of the six
benchmark--divergence pairs (Table~4 of the main paper).

\section{Experimental Setup}
\label{sec:supp_visual_experiments}

\subsection{Visual On-Policy Self-Distillation Baselines}
\label{sec:supp_visual_baselines}

In all three methods, the student samples a response $\hat y$ given
the full image $x$ and question $q$. The methods differ in how they
use visual evidence to supervise the next token at the student's
prefix $\hat y_{<t}$.

\paragraph{Vision-OPD.}
The teacher receives a local crop $x'$ containing the relevant
detail. At the same question and prefix it produces a next-token
distribution $p_T(\cdot\mid\hat y_{<t},x',q)$, which directly
supervises the full-image student~\cite{yuan2026vision}.

\paragraph{OPD-V.}
The teacher predicts under both zoomed-in and masked views. OPD-V
uses the contrast between these predictions to select and weight
generation positions~\cite{bi2026opd}. This changes how much a given
prefix contributes to the training loss.

\paragraph{VAD.}
VAD uses visual evidence attribution to reconstruct the distribution
that supervises the student~\cite{zhang2026vad}. This changes the
target probabilities over candidate tokens at a given prefix.

\subsection{Integrating ADD with the Baselines}
\label{sec:supp_add_integration}

At each student-generated prefix, ADD compares the baseline's target
distribution with the full-image student's next-token prediction.
For Vision-OPD, the target is the crop-conditioned prediction.
OPD-V retains its selection and weighting of generation positions,
and ADD weights candidate regions within each selected position.
For VAD, ADD uses the reconstructed target both to form the regions
and to supervise the student's relative probabilities within them.

Our main comparisons use reverse KL for each baseline and its ADD
variant. Following Vision-OPD~\cite{yuan2026vision}, we retain the
student's $K=100$ most probable tokens at each prefix, denoted by
$\mathcal C_K$, and take the teacher probabilities on the same set.
For $a\in\{T,S\}$, the retained tokens have total mass
$w_K^a=\sum_{v\in\mathcal C_K}p_a(v)$ and conditional probabilities
$p_a^K(v)=p_a(v)/w_K^a$. All other tokens form one outcome with
mass $w_{\mathrm{tail}}^a=1-w_K^a$. For the RKL variant of
Vision-OPD, substituting $p_a(v)=w_K^a p_a^K(v)$ into the loss
on these $K+1$ outcomes gives
\begin{equation}
\begin{aligned}
 D_{\mathrm{RKL}}^{(K)}
 &= \sum_{v\in\mathcal C_K}p_S(v)\log\frac{p_S(v)}{p_T(v)}
    +w_{\mathrm{tail}}^S\log\frac{w_{\mathrm{tail}}^S}{w_{\mathrm{tail}}^T}\\
 &= \sum_{R\in\{K,\mathrm{tail}\}}w_R^S\log\frac{w_R^S}{w_R^T}
    +w_K^S\,\mathrm{KL}(p_S^K\|p_T^K).
\end{aligned}
\label{eq:supp_topk_tail_rkl}
\end{equation}
The first term on the second line compares the mass assigned to
$\mathcal C_K$ and to the tail. The second aligns the relative
probabilities of the retained tokens, weighted by their total
student mass $w_K^S$. The tail enters only through its total mass.

Within $\mathcal C_K$, let $\mathcal V^+$ contain the tokens with
$p_T(v)\geq p_S(v)$ and $\mathcal V^-$ the rest. With
$\boldsymbol w^a=(w_+^a,w_-^a,w_{\mathrm{tail}}^a)$, regrouping the
token sum in Eq.~\eqref{eq:supp_topk_tail_rkl} gives the same baseline loss,
\begin{equation}
 D_{\mathrm{RKL}}^{(K)}
 =\underbrace{\mathrm{KL}(\boldsymbol w^S\|\boldsymbol w^T)}_{\mathcal L_{\mathrm{mass}}}
  +\sum_{R\in\{+,-\}}w_R^S D_{\mathrm{RKL}}^R .
\label{eq:supp_topk_sign_rkl}
\end{equation}
Here $D_{\mathrm{RKL}}^R=\mathrm{KL}(p_S^R\|p_T^R)$ compares relative
token probabilities within region $R$. ADD keeps $\mathcal L_{\mathrm{mass}}$
and replaces $w_R^S$ in the conditional term with
$\lambda\,\operatorname{sg}(w_R^*)$, where
$w_R^*=\max\{w_R^T,w_R^S\}$ and $\lambda=0.7$ in the main comparisons.
We recompute $w_R^*$ at each prefix and hold it fixed during
backpropagation. Both masses come from the baseline predictions, so
ADD requires no additional forward pass.

\subsection{Datasets}
\label{sec:supp_visual_datasets}

\subsubsection{Visual training data and setup}
\begin{table}[!t]
\centering
\caption{Visual prompts. In the training template, \texttt{<image>} denotes the boxed full image for the student and the crop for the Vision-OPD teacher. In the evaluation templates, it denotes the full benchmark image.}
\label{tab:supp_prompt_templates}
\begingroup
\newcommand{\promptblock}[2]{%
  \noindent\begin{minipage}[t]{\linewidth}
    \setlength{\fboxsep}{3pt}%
    \colorbox{black}{\parbox{\dimexpr\linewidth-2\fboxsep\relax}{%
      \normalfont\small\bfseries\color{white}\strut #1}}\par
    \vspace{2pt}\noindent\hspace*{\fboxsep}%
    \begin{minipage}[t]{\dimexpr\linewidth-2\fboxsep\relax}
      \raggedright\footnotesize\ttfamily #2\strut
    \end{minipage}\par
  \end{minipage}\par\smallskip}
\newcommand{\imgslot}{\textless image\textgreater}
\newcommand{\qslot}{\{question\}}
\newcommand{\optslot}{\{options\}}
\promptblock{Vision-OPD training}{%
  \imgslot\\
  \qslot\\[2pt]
  Only focus on the objects inside the red bounding box in the image to answer this question.}
\promptblock{V$^*$Bench}{%
  \imgslot\\
  \qslot\\
  \optslot\\
  Answer with the option's letter from the given choices directly.}
\promptblock{ZoomBench}{%
  \imgslot\\
  \qslot\\
  \optslot}
\promptblock{HR-Bench 4K and 8K}{%
  \imgslot\\
  \qslot{} Select from the following choices.\\
  \optslot}
\promptblock{MME-RealWorld EN}{%
  \imgslot\\
  \qslot{} The choices are listed below:\\
  \optslot\\
  Select the best answer to the above multiple-choice question based on the image. Respond with only the letter (A, B, C, D, or E) of the correct option.\\
  The best answer is:}
\promptblock{MME-RealWorld CN}{%
  \imgslot\\
  \qslot{} \promptchinese{选项如下所示：}\\
  \optslot\\
  \promptchinese{根据图像选择上述多项选择题的最佳答案。只需回答正确选项的字母（A, B, C, D 或 E）。}\\
  \promptchinese{最佳答案为：}}
\endgroup
\end{table}

We train Qwen3.5-2B and Qwen3.5-4B for one epoch on the 6,241
synthetic visual question--answer pairs released by
Vision-OPD~\cite{qwen3.5,yuan2026vision}. Each example supplies the
question, full image, and local crop. The full-image student samples
the response, so the prefixes used for distillation follow the
student's own predictions. The maximum prompt and response lengths
are 8,192 and 1,024 tokens, respectively.

Within each model size, we compare each baseline with its ADD
variant under matched training hyperparameters and evaluate the
final checkpoint. Vision-OPD and OPD-V update the teacher by an
exponential moving average of the student with rate $0.05$.
VAD uses a frozen teacher. Training uses NVIDIA RTX PRO 6000 GPUs.

\paragraph{Training and evaluation prompts.}
We follow the training format of Vision-OPD~\cite{yuan2026vision}.
The student sees the full image with a red box around the target.
The teacher receives the corresponding crop in the image slot and
scores the next token at the same student-generated prefix
$\hat y_{<t}$. Table~\ref{tab:supp_prompt_templates} shows the
training prompt and the evaluation prompts from Appendix~B of
Vision-OPD. At test time, we pass the full benchmark image to the
student and insert the question and choices in the language of
the benchmark. Both HR-Bench resolutions use the same template.

\subsubsection{Visual evaluation data}
The main visual comparison evaluates the full-image student in six
settings from four benchmark families. Table~\ref{tab:supp_benchmark_sizes}
gives the number of examples in each setting. The released Vision-OPD
pipeline parses answer choices from generated responses and queries
DeepSeek V4.1 Flash as a fallback judge when the parser returns no
valid choice~\cite{yuan2026vision,xu2026deepseek}.

\par\medskip\noindent\begin{minipage}{\linewidth}
\captionsetup{type=table}
\centering
\caption{Evaluation samples in each visual benchmark setting.}
\label{tab:supp_benchmark_sizes}
\small
\begin{tabular}{@{}lr@{}}
\toprule
Benchmark & Samples \\
\midrule
V$^*$Bench       & 191 \\
ZoomBench        & 845 \\
HR-Bench 4K      & 800 \\
HR-Bench 8K      & 800 \\
MME-RealWorld EN & 23,609 \\
MME-RealWorld CN & 5,917 \\
\bottomrule
\end{tabular}
\end{minipage}\par\medskip

\paragraph{V$^*$Bench and ZoomBench.}
V$^*$Bench tests attribute recognition and spatial relations in
cluttered images~\cite{wu2024v}. ZoomBench asks about details that
occupy a small region, including text, material, and object
structure~\cite{wei2026zooming}. Both datasets also provide target
regions. In the conditional-alignment analysis, these regions
provide a privileged crop for the teacher. The teacher and initial
student define a common partition on which the trained models
are compared.

\paragraph{HR-Bench and MME-RealWorld.}
HR-Bench tests whether the model can resolve details of individual
instances and combine evidence across instances. We report its 4K
and 8K settings separately~\cite{wang2025divide}. MME-RealWorld
contributes English and Chinese evaluations spanning scene text,
chart interpretation, remote sensing, and autonomous
driving~\cite{zhang2025mme}.

\paragraph{Averaging across settings.}
The main comparison averages all six accuracies. The $\lambda$
sweep and fixed-weight comparison average five settings, using
V$^*$Bench, ZoomBench, HR-Bench 4K and 8K, and MME-RealWorld CN.
In the component and weight-granularity ablations, HR-Bench Avg.\
is the mean of the 4K and 8K accuracies.

\subsubsection{Evaluation beyond visual distillation}
For the text-only experiments, Qwen3-4B-Instruct-2507 teaches
Qwen3-1.7B and Qwen3-0.6B. Standard OPD minimizes reverse KL
(RKL), $\mathrm{KL}(p_S\|p_T)$, at each student-generated prefix.
ADD uses the same RKL direction within each region.

For mathematical reasoning, we use DeepMath~\cite{he2026deepmath} problems with a
difficulty level of at least $6$, yielding approximately 57K
training prompts. For code generation, we use Eurus-RL-Code~\cite{cui2025process},
which contains approximately 25K training prompts.
We use the task-specific prompt templates in
Table~\ref{tab:supp_text_prompt_templates} for training.

\begin{table}[!t]
\centering
\caption{Training and evaluation prompts for mathematical
reasoning and code generation.
\texttt{\{question\}} denotes the problem statement.}
\label{tab:supp_text_prompt_templates}
\begingroup
\newcommand{\promptblock}[2]{%
  \noindent\begin{minipage}[t]{\linewidth}
    \setlength{\fboxsep}{3pt}%
    \colorbox{black}{%
      \parbox{\dimexpr\linewidth-2\fboxsep\relax}{%
        \normalfont\small\bfseries\color{white}\strut #1}}\par
    \vspace{2pt}\noindent\hspace*{\fboxsep}%
    \begin{minipage}[t]{\dimexpr\linewidth-2\fboxsep\relax}
      \raggedright\footnotesize\ttfamily #2\strut
    \end{minipage}\par
  \end{minipage}\par\smallskip}
\newcommand{\qslot}{\{question\}}
\newcommand{\imstart}{\textless|im\_start|\textgreater}
\newcommand{\imend}{\textless|im\_end|\textgreater}
\newcommand{\codefence}{\char96\char96\char96}

\promptblock{Mathematical reasoning}{%
  \imstart{}user\\
  \qslot\\
  Please reason step by step, and put your final answer
  within \textbackslash boxed\{\}.\imend\\
  \imstart{}assistant}

\promptblock{Code generation}{%
  \imstart{}user\\
  \qslot\\
  Write Python code to solve the problem. Present the code in\\
  \codefence{}python\\
  Your code\\
  \codefence\\
  at the end.\\
  You need to think first then write the Python code.\imend\\
  \imstart{}assistant}

\endgroup
\end{table}

For both student sizes, the base learning rate is $5\times10^{-6}$.
We keep it constant for mathematics and decay it linearly for code,
with zero warmup steps in both cases. We train LoRA adapters of
rank $64$ with $\alpha=128$ and dropout $0.05$. Training uses
bfloat16 and gradient checkpointing. The student samples training
responses at temperature $0.7$.

We evaluate mathematical reasoning on AIME24, AIME25, and
HMMT25~\cite{balunovic2026matharena}, and code generation on HE+
and MBPP+~\cite{liu2023your}. For each problem, we average
correctness over 16 independently sampled responses for mathematics
(mean@16) and four for code (mean@4). Each benchmark has equal
weight in the average for its task.

\section{More Experiments}
\label{sec:supp_additional_ablations}

We use the Qwen3.5-4B Vision-OPD setup from the main paper. Mean
accuracy gives equal weight to the five evaluation settings, counting
HR-Bench 4K and 8K separately.

\subsection{Sensitivity to the Top-\texorpdfstring{$K$}{K} Cutoff}
\label{sec:supp_topk}

We use $K=100$ in the main comparisons to match the candidate cutoff
of the baselines. To measure sensitivity to this choice, we vary $K$
from 20 to 200 (Table~\ref{tab:supp_topk}). Increasing $K$ brings
lower-ranked tokens into the positive and negative regions, where
their relative probabilities receive supervision. Tokens outside the
student's top $K$ contribute only through their combined mass in the
tail bucket.

\begin{table}[htbp]
\centering
\caption{Accuracy (\%) by size of the student's top-$K$ set.}
\label{tab:supp_topk}
\footnotesize
\renewcommand{\arraystretch}{1.12}
\setlength{\tabcolsep}{3pt}
\begin{tabular*}{\linewidth}{@{\extracolsep{\fill}}lcccc@{}}
\toprule
Benchmark & $K=20$ & $K=50$ & $K=100$ & $K=200$ \\
\midrule
V$^*$Bench       & 86.91 & 90.05 & \textbf{92.67} & 92.15 \\
ZoomBench        & 55.98 & 58.82 & 60.71 & \textbf{60.83} \\
HR-Bench 4K      & 81.50 & 82.00 & 82.38 & \textbf{83.50} \\
HR-Bench 8K      & 80.63 & 81.63 & 81.88 & \textbf{82.13} \\
MME-RealWorld CN & 69.41 & 69.85 & \textbf{70.53} & 70.26 \\
\midrule
Avg.             & 74.89 & 76.47 & 77.63 & \textbf{77.77} \\
\bottomrule
\end{tabular*}
\end{table}

Increasing $K$ from 20 to 100 improves all five settings and raises
mean accuracy from 74.89\% to 77.63\%. Doubling $K$ to 200 adds 0.14
points on average, with a further 1.12-point gain on HR-Bench 4K and
a 0.52-point drop on V$^*$Bench. Thus the baseline cutoff captures
most of the observed gain from expanding the candidate set.

\subsection{Adaptive Splits at Fixed Total Weight}
\label{sec:supp_equal_budget}

Under RKL, a student that assigns too little mass to teacher-favored
candidates also gives their conditional error too little weight. ADD
restores the positive-region coefficient to the teacher mass
$w_{+,t}^*=w_{+,t}^T$, while retaining the original negative-region
coefficient $w_{-,t}^*=w_{-,t}^S$. The fixed-weight and sequence-mean
ablations in the main paper can change both the sum and the ratio of
the two coefficients at a given prefix. Here the control recomputes
the ADD weights at every prefix and assigns their mean to each
region,
\begin{equation}
 w_{+,t}^{\mathrm{eq}} = w_{-,t}^{\mathrm{eq}}
 = \frac{w_{+,t}^*+w_{-,t}^*}{2}.
 \label{eq:supp_region_weights}
\end{equation}

For given teacher and student predictions,
Eq.~\eqref{eq:supp_region_weights} preserves the sum of the
coefficients. The weighted conditional loss and its gradient still
depend on which coefficient multiplies each regional error. For
example, if ADD assigns 0.8 to the positive region and 0.4 to the
negative region, averaging gives each 0.6. The positive coefficient then falls 0.2 short of the teacher mass that ADD restores. At
prefixes where $w_{-,t}^S>w_{+,t}^T$, averaging instead cuts the
negative-region weight below the value RKL already assigns it. Both
runs use the same mass-calibration term, $\lambda=0.7$, and
stop-gradient on the weights.

Equal splitting gives 75.80\% mean accuracy, below Vision-OPD at
76.61\% (Table~\ref{tab:supp_equal_budget}). ADD reaches 77.63\% and
improves four of the five settings over equal splitting. The
1.83-point difference supports restoring the coefficient suppressed
by RKL while preserving the weight on the other region, even when the
total coefficient already adapts to each prefix.

\begin{table}[htbp]
\centering
\caption{Accuracy (\%) under two allocations of the same sum of
conditional coefficients. Vision-OPD is the 4B baseline from the main paper.}
\label{tab:supp_equal_budget}
\footnotesize
\renewcommand{\arraystretch}{1.12}
\setlength{\tabcolsep}{2pt}
\begin{tabular*}{\linewidth}{@{\extracolsep{\fill}}lccc@{}}
\toprule
Benchmark & Vision-OPD & Equal split & ADD \\
\midrule
V$^*$Bench       & 89.53 & 89.53 & \textbf{92.67} \\
ZoomBench        & 59.88 & 56.45 & \textbf{60.71} \\
HR-Bench 4K      & \textbf{83.00} & 82.62 & 82.38 \\
HR-Bench 8K      & 80.13 & 80.75 & \textbf{81.88} \\
MME-RealWorld CN & 70.49 & 69.66 & \textbf{70.53} \\
\midrule
Avg.             & 76.61 & 75.80 & \textbf{77.63} \\
\bottomrule
\end{tabular*}
\end{table}

\subsection{Targeting High-Disagreement Positions}
\label{sec:supp_position_selection}

At which positions does restoring regional supervision matter most?
Over the full vocabulary, $w_+^*=w_+^T$ and $w_-^*=w_-^S$. Since
$w_+^S+w_-^S=1$, the two conditional coefficients sum to
\[
 w_+^*+w_-^*
 =1+(w_+^T-w_+^S)
 =1+\tfrac12\lVert p_T-p_S\rVert_1 .
\]
A larger disagreement therefore increases the total coefficient,
while the masses $w_+^T$ and $w_-^S$ determine its division between
regions. In the top-$K$ implementation, the corresponding sum is
$w_K^S+(w_+^T-w_+^S)$, where $w_K^S=1-w_{\mathrm{tail}}^S$ and the
two positive masses are taken over retained candidates. We use the
full-vocabulary distance $d_{\ell_1}=\lVert p_T-p_S\rVert_1$ to rank
positions, as in Fig.~1 of the main paper.

Every valid position contributes to distillation. Within each
micro-batch, we use the complete ADD rule at the highest-disagreement
25\% of positions or at an equal number sampled uniformly without
replacement. All remaining positions use the original RKL weights.
Vision-OPD and full ADD use their respective weights at every
position.

\begin{table}[htbp]
\centering
\caption{Accuracy (\%) by choice of positions receiving ADD. V$^*$Bench,
ZoomBench, and MME-RealWorld CN contain 191, 845, and 5,917
questions. HR-Bench Avg.\ averages the 4K and 8K splits (800
questions each).}
\label{tab:supp_position_selection}
\footnotesize
\setlength{\tabcolsep}{1.5pt}
\renewcommand{\arraystretch}{1.12}
\begin{tabular*}{\linewidth}{@{}l@{\extracolsep{\fill}}cccc@{}}
\toprule
Position selection & V$^*$ & ZoomBench &
\begin{tabular}[c]{@{}c@{}}HR-Bench\\Avg.\end{tabular} &
\begin{tabular}[c]{@{}c@{}}MME-RealWorld\\CN\end{tabular} \\
\midrule
Vision-OPD & 89.53 & 59.88 & 81.57 & 70.49 \\
Random 25\% & 89.01 & 60.12 & 81.57 & 70.44 \\
Top 25\% by $d_{\ell_1}$ & 91.62 & 60.47 & \textbf{82.19} & 70.51 \\
ADD (all) & \textbf{92.67} & \textbf{60.71} & 82.13 & \textbf{70.53} \\
\bottomrule
\end{tabular*}
\end{table}

At the same 25\% coverage, random selection gives 76.54\% mean
accuracy, close to Vision-OPD's 76.61\%. Selecting by disagreement
gives 77.40\% (Table~\ref{tab:supp_position_selection}). Random
selection reaches only about a quarter of the high-disagreement
positions on average. Ranking reaches all of them, so ADD restores
supervision where Fig.~1(c) of the main paper finds larger positive-region errors
paired with smaller RKL weights. This offers an explanation for the
0.86-point gain over random selection.

The high-disagreement quarter improves on Vision-OPD in all five
settings and recovers about three quarters of full ADD's mean gain.
This comparison locates much of the benefit of the complete weighting
rule at high-disagreement positions.
Section~\ref{sec:supp_equal_budget} separately tests how the two
regions share a given total coefficient. Full ADD still has the
highest mean (77.63\%) and reaches 92.67\% on V$^*$Bench, compared
with 91.62\% for the high-disagreement quarter.
\begin{figure*}[tp]
\centering
\includegraphics[width=\textwidth]{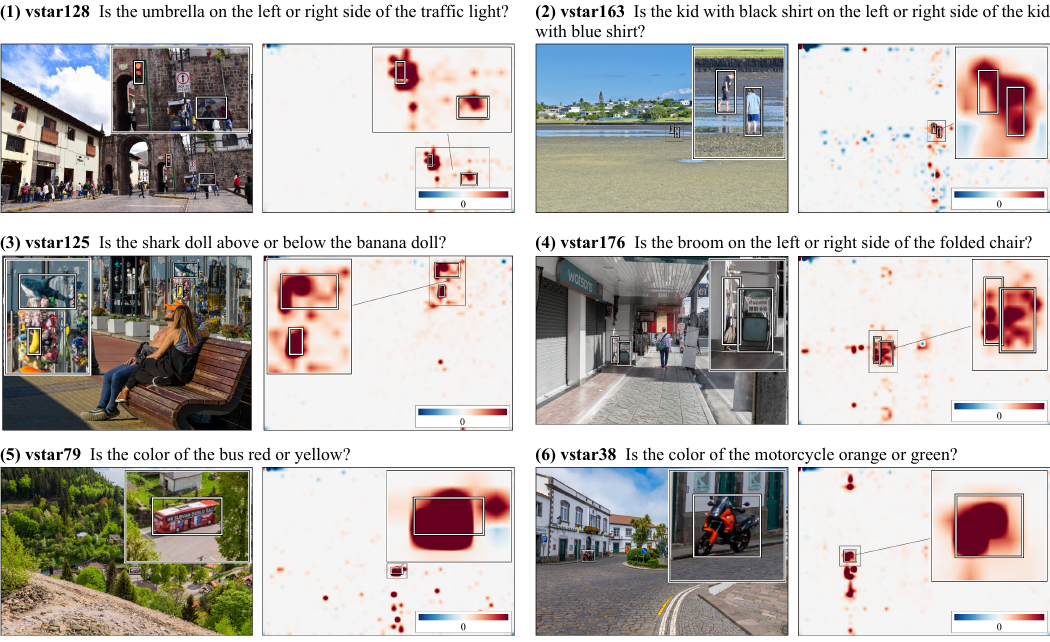}
\caption{Response-averaged attention differences on V$^*$Bench.
Each pair shows the input image (left) and the difference on the
image-token grid (right), computed as Vision-OPD+ADD minus Vision-OPD.
Red indicates higher attention with ADD and blue indicates lower
attention. Each map has its own zero-centered color scale. Insets
enlarge the boxed regions.}
\label{fig:supp_attention}
\end{figure*}
\subsection{The Optimization Shortcut}
\label{sec:supp_shortcut_step}

Under coupled RKL, a student can give a regional error less weight
simply by assigning less probability to that region. To see whether
this occurs during an update, we randomly sampled 32 questions and
generated eight responses per question with the initial Qwen3.5-4B
student. We used the original RKL gradient on these 256 sequences to
update all student parameters once. Replaying the same prefixes with
the EMA teacher distribution and the region partition fixed lets us
compare the mass and the conditional error for the same candidates
before and after the update.

In \textbf{14,118 of 61,158} prefix--region pairs \textbf{(23.08\%)},
student mass fell ($\Delta w_R^S<0$) while unweighted conditional RKL
held steady or rose ($\Delta D_{\mathrm{RKL}}^R\geq 0$). For these
pairs the student put less weight on an error it had not corrected.
Any decrease in their weighted penalty $w_R^S D_{\mathrm{RKL}}^R$
may therefore stem from a reduction in weight rather than
improved within-region alignment. This is
consistent with the shortcut identified in
Eq.~\eqref{eq:supp_rkl_gradient} and explains why ADD holds the
coefficient fixed when differentiating the conditional loss.

\subsection{Attention Differences on \texorpdfstring{V$^*$Bench}{V*Bench}}
\label{sec:supp_attention}

Figure~1(a) of the main paper relates teacher--student disagreement
to the total attention paid to image tokens. To see where attention
changes with ADD, we compare Vision-OPD and Vision-OPD+ADD on six
further V$^*$Bench questions (Supplementary Fig.~\ref{fig:supp_attention}). For
each image, the models generate independent responses to the same
question. We average attention to each image token over the eight
full-attention layers and 16 query heads, then over the generated
tokens in each model's own response.

Judging whether the umbrella is to the left or right of the traffic
light requires locating both objects. In (1), both have higher
response-averaged attention with ADD. For the color questions in (5)
and (6), the increase overlaps the bus and the motorcycle,
respectively, even though each occupies a small part of the scene.
These maps illustrate how the change in full-image attention can
coincide with the local evidence needed for a spatial relation or an
attribute.

\section{More Discussion}
\label{sec:supp_more_discussion}
\subsection{Why Split by Sign?}
\label{sec:supp_partition_rationale}

At a given prefix, the student can assign too little probability to a
set of tokens and also get their relative probabilities wrong. To see
why the first error can mask supervision for the second, consider the
tokens in $\mathcal V^+=\{v:p_T(v)\geq p_S(v)\}$. Every probability
shortfall lies in this set, and every excess lies in $\mathcal V^-$.
Over the full vocabulary, the positive region therefore captures the
largest possible mass shortfall of any candidate set,
\[
 w_+^T-w_+^S
 = \max_{G\subseteq\mathcal V}\sum_{v\in G}\bigl(p_T(v)-p_S(v)\bigr)
 = \tfrac12 d_{\ell_1},
\]
where $d_{\ell_1}=\lVert p_T-p_S\rVert_1$. Thus, at positions with
greater disagreement, the student falls further short of the teacher
in the mass it assigns to these candidates.

The choice of set matters. A teacher-ranked head~\cite{wu2025rethinking}
can contain tokens the student underestimates alongside others it
overestimates. Probability missing from the former can reappear on the
latter without changing the head mass at all. The same cancellation
can occur among the non-target alternatives in a target/non-target
split~\cite{zhao2022decoupled}, or among tokens shared by the teacher's
and student's top-$K$ sets~\cite{li2026rethinking}. The RKL
coefficient for such a group can therefore stay fixed even as the
student assigns less mass to its underestimated tokens. Splitting by
sign keeps that offsetting mass in a separate region.

For the very candidates whose mass falls short, $D_{\mathrm{RKL}}^+$
measures how far the student's relative probabilities are from the
teacher's. Reverse KL scales the gradient that corrects those
proportions by $w_+^S$ (Eq.~\eqref{eq:supp_rkl_gradient}).
Figure~1(c) of the main paper compares these two quantities across
generation positions. As $d_{\ell_1}$ increases, the unweighted
$D_{\mathrm{RKL}}^+$ generally rises, while $w_+^S$ itself generally
falls. The sign split makes supervision masking visible because both
measurements track the same candidates at each prefix. Where the
student makes larger errors in their relative probabilities, RKL
typically gives their correction less weight.

\subsection{From Group DRO to Regional Min--Max}
\label{sec:supp_minmax_motivation}

A model can incur a large mean loss on a rare group while achieving a
low loss overall. The group's frequency determines how heavily its
errors count in the average. Group DRO allows the group
proportions to vary and minimizes the loss under the worst admissible
mixture~\cite{sagawa2019distributionally}.

At a single generation position, the student's own prediction can hide
an error in much the same way. The student mass $w_R^S$ plays the role
of group frequency in reverse KL. It weights the error
$D_{\mathrm{RKL}}^R$ in the relative probabilities within region $R$.
Consider $\mathcal V^+$, where the student assigns less mass than the
teacher. Scaling up all student probabilities in this region by the
same factor can bring its mass to $w_+^T$ while leaving the relative
probabilities unchanged. The same error would then contribute
$w_+^T D_{\mathrm{RKL}}^+$ to the loss, although the current objective
counts it only at $w_+^S D_{\mathrm{RKL}}^+$.

We borrow Group DRO's worst-case criterion to choose the weight for
each region. How large could the regional penalty be as its mass moves
from $w_R^S$ toward $w_R^T$? In Eq.~(9) of the main paper, the worst
case occurs when the current conditional error persists at the larger
of these two masses. We take this maximum separately in each region
and minimize the sum of the
resulting penalties. Thus $w_R^*=\max\{w_R^S,w_R^T\}$ restores the
teacher's weight on underestimated candidates and preserves the
student's weight on overestimated ones. We hold each coefficient fixed
during the update so that reducing its conditional penalty requires
better alignment within that region.


\end{document}